\pdfoutput=1

\documentclass[11pt]{article}

\usepackage{acl}

\usepackage{times}
\usepackage{latexsym}

\usepackage[T1]{fontenc}

\usepackage[utf8]{inputenc}

\usepackage{microtype}

\usepackage{inconsolata}

\usepackage{amsmath,amsfonts,bm}

\def\eqref#1{equation~\ref{#1}}

\def\1{\bm{1}}

\def\vb{{\bm{b}}}

\def\vg{{\bm{g}}}

\def\vx{{\bm{x}}}
\def\vy{{\bm{y}}}

\def\mA{{\bm{A}}}
\def\mB{{\bm{B}}}

\def\mG{{\bm{G}}}

\def\mI{{\bm{I}}}

\def\mW{{\bm{W}}}

\DeclareMathAlphabet{\mathsfit}{\encodingdefault}{\sfdefault}{m}{sl}
\SetMathAlphabet{\mathsfit}{bold}{\encodingdefault}{\sfdefault}{bx}{n}

\usepackage{graphicx}
\usepackage{amsmath}
\usepackage{amssymb}
\usepackage{mathtools}
\usepackage{amsthm}
\usepackage{bm}
\usepackage{sidecap}
\usepackage[most]{tcolorbox}  

\usepackage{subfigure}
\usepackage{booktabs} 
\usepackage{colortbl}
\usepackage{multirow}
\usepackage{bbm}
\usepackage{enumitem}
\usepackage{makecell} 

\setenumerate[1]{itemsep=0pt, partopsep=0pt, parsep=\parskip, topsep=5pt}
\setitemize[1]{itemsep=0pt, partopsep=0pt, parsep=\parskip, topsep=5pt}
\setdescription[1]{itemsep=0pt, partopsep=0pt, parsep=\parskip, topsep=5pt}

\usepackage[capitalize,noabbrev]{cleveref}

\usepackage{subcaption}
\usepackage{subfloat}

\definecolor{darkgrey}{rgb}{0.53,0.53,0.53}
\definecolor{mygrey}{rgb}{0.9,0.9,0.9}
\definecolor{purple}{RGB}{230, 227, 254}
\definecolor{lightgreen}{RGB}{238, 252, 241}
\definecolor{lightred}{RGB}{231, 187, 187}
\definecolor{darkred}{RGB}{198, 129, 129}

\definecolor{tabhighlight}{HTML}{e5e5e5}
\definecolor{someorange}{rgb}{0.773,0.353,0.067}
\definecolor{someblue}{rgb}{0.27, 0.35, 0.760}

\theoremstyle{plain}
\newtheorem{theorem}{Theorem}[section]

\newtheorem{property}[theorem]{Property}

\theoremstyle{definition}

\theoremstyle{remark}

\usepackage[textsize=tiny]{todonotes}

\title{Each Rank Could be an Expert: Single-Ranked Mixture of Experts LoRA for Multi-task Learning}

\author{
Ziyu Zhao\textsuperscript{1}\textsuperscript{2}\thanks{~~Equal Contribution.},
Yixiao Zhou\textsuperscript{1}\textsuperscript{2}\footnotemark[1],
Zhi Zhang\textsuperscript{3},
Didi Zhu\textsuperscript{1},
Tao Shen\textsuperscript{1},
Zexi Li\textsuperscript{1},
Jinluan Yang\textsuperscript{1}, \\
\textbf{Xuwu Wang}\textsuperscript{3},
\textbf{Jing Su}\textsuperscript{3}, 
\textbf{Kun Kuang}\textsuperscript{1}\textsuperscript{5},
\textbf{Zhongyu Wei}\textsuperscript{4},
\textbf{Fei Wu}\textsuperscript{1} 
\textbf{Yu Cheng}\textsuperscript{5} \\
\textsuperscript{1}Zhejiang University,
\textsuperscript{2}Shanghai Innovation Institute,
\textsuperscript{3}ByteDance Inc.,\\
\textsuperscript{4}Fudan University,
\textsuperscript{5}The Chinese University of Hong Kong
}

\begin{document}
\maketitle
\begin{abstract}
Low-Rank Adaptation (LoRA) is widely used for adapting large language models (LLMs) to specific domains due to its efficiency and modularity. However, vanilla LoRA struggles with task conflicts in multi-task scenarios. Recent works use a Mixture of Experts (MoE) approach, treating each LoRA module as a specialized expert to reduce interference. However, these MoE methods often isolate knowledge within individual tasks, failing to leverage shared knowledge across related tasks. In this paper, we present a unified framework for single LoRA and multi-LoRA MoE. Within this framework, we propose Single-ranked Mixture of Experts LoRA (\textbf{SMoRA}), which embeds MoE into a singular LoRA by \textit{treating each rank as an independent expert}. With a \textit{\textbf{dynamic rank-wise sparse activation}} mechanism, SMoRA achieves an automatic rank dispatching that promotes finer-grained knowledge sharing while mitigating task conflicts. Experiments demonstrate that SMoRA activates fewer parameters yet achieves better performance in multi-task scenarios.
\end{abstract}

\begin{figure*}
    \centering
\includegraphics[width=.85\linewidth]{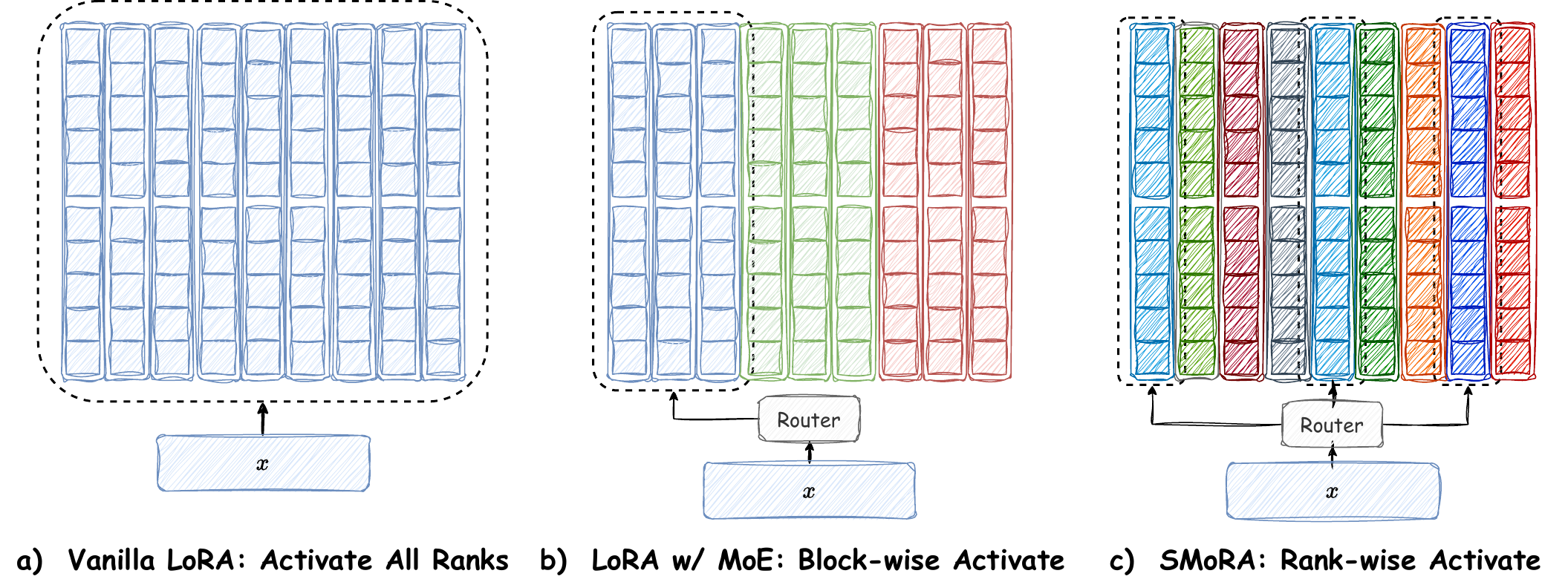}
    \caption{\textbf{Comparison of Vanilla LoRA, LoRA MoE, and SMoRA}: (a) Vanilla LoRA uniformly activates all ranks, lacking a mechanism to differentiate parameters for distinct tasks, which degrades performance on heterogeneous downstream tasks. (b) LoRA MoE addresses task interference by explicitly partitioning LoRA parameters. However, it fails to achieve finer-grained parameter combinations and effective knowledge sharing, leading to suboptimal performance. (c) SMoRA views each rank as an expert, incorporating MoE within LoRA to enable dynamic rank activation. This design balances task-specific parameter customization and knowledge sharing effectively.}
    \label{fig:compare}
    \vspace{-4mm}
\end{figure*}

With the remarkable success of LLMs across various domains~\cite{dubey2024llama,liu2023gpt}, there has been growing interest in fine-tuning them for diverse, task-specific applications~\cite{zhang2023instruction}. However, full-parameter fine-tuning is often computationally prohibitive, prompting the adoption of Parameter-Efficient Fine-Tuning (PEFT) approaches~\cite{han2024parameter}. Among these, LoRA has gained prominence for its efficiency and modular design~\cite{han2024parameter,zhao2024loraretriever}.
Despite its advantages, LoRA encounters a critical limitation in multi-task settings: it uniformly activates all parameters during fine-tuning, without decoupling them to adapt to the specific needs of heterogeneous downstream tasks, as shown in Fig.\ref{fig:compare}a). This lack of task-specific adaptation often leads to performance degradation in multi-task scenarios~\cite{tian2024hydralora,liu2023moelora,zhao2024merging,feng2024mixture}.

To address this challenge, recent studies have incorporated the MoE framework into LoRA training~\cite{mao2025survey,dou2023loramoe,liu2023moelora,zadouri2023pushing}. 
As illustrated in Fig.\ref{fig:compare}b), these approaches construct multiple LoRAs, treating each module as an independent expert. A router is then employed to dynamically activate specific modules, enhancing task adaptation and computational efficiency~\cite{zhu2023sira,li2024mixlora,feng2024mixture}.
However, these methods partition the parameter space into several fixed submatrices, where each block is activated or deactivated as a whole. This block-wise activation is inherently coarse-grained, which fails to capture subtle distinctions or shared knowledge across tasks, leading to inefficient utilization of learned knowledge. Furthermore, the rigid division of modules restricts the dynamic combination and decomposition of knowledge, ultimately limiting the model's expressive power and adaptability in multi-task learning scenarios.

\begin{figure*}
    \centering
\includegraphics[width=\linewidth]{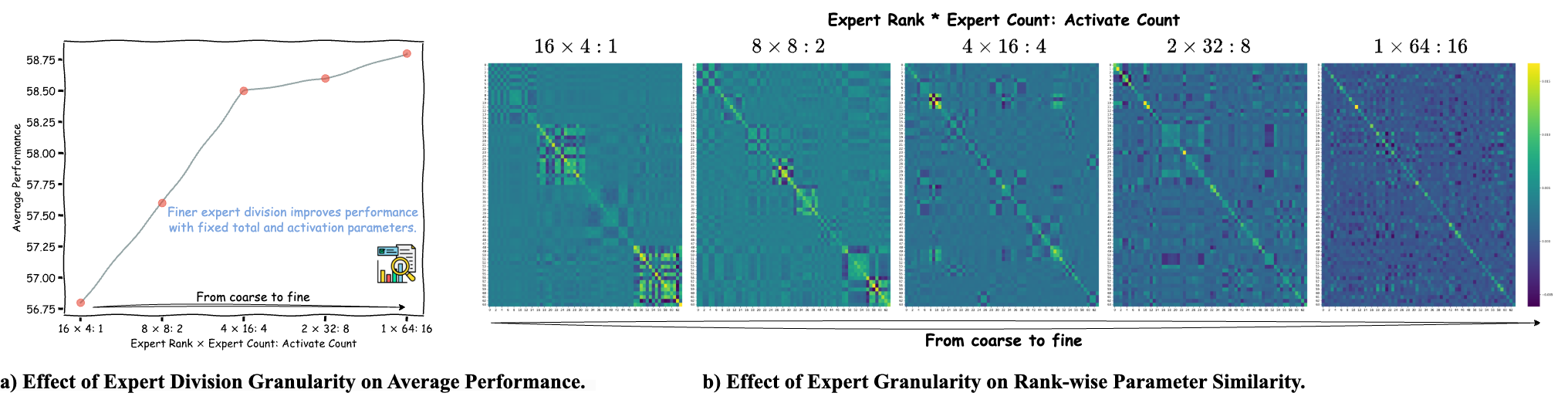}
    \caption{\textbf{Varying Expert Granularity.} Effects of altering expert granularity while keeping total rank capacity (Expert Rank × Expert Count = 64) and activated rank (Expert Rank × Activate Count = 16) constant. a) Average performance demonstrates an increase as expert granularity becomes finer. b) Rank-wise parameter similarity shifts from block-diagonal structures (coarse-grained) to strong main-diagonal patterns (fine-grained), implying more distinct fine-grained experts capable of representing a larger, more diverse parameter space.
    }
    \label{fig:motivation}
    \vspace{-4mm}
\end{figure*}

In this paper, we first provide a unifying perspective, demonstrating that multi-LoRA MoE can be represented as a single LoRA, where the rank is partitioned into blocks and each block is activated independently (see~\S\ref{sec:equivalence} for details). 
Through extensive experiments, we further investigate the impact of block granularity on downstream performance under fixed total and activated parameter budgets. Interestingly, we observe that \textit{finer-grained partitioning of the rank space yields significantly better performance and allows the resulting learned experts to represent a larger and more diverse parameter space}, as illustrated in Fig.~\ref{fig:motivation} (see \S\ref{subsec:motivation} for details).
Based on these findings, we propose the following key insight:
\begin{tcolorbox}[notitle, rounded corners, colframe=darkgrey, colback=white, boxrule=2pt, boxsep=0pt, left=0.15cm, right=0.17cm, enhanced, shadow={2.5pt}{-2.5pt}{0pt}{opacity=5,mygrey},toprule=2pt, before skip=0.65em, after skip=0.75em 
  ]
\emph{
  {
    \centering 
  {
    \fontsize{8pt}{13.2pt}\selectfont 
Each rank of LoRA can be treated as an expert to achieve finer-grained knowledge sharing and adaptation.
}  \\
  }
  }
\end{tcolorbox}


Building on this motivation, we propose \textbf{S}ingle-ranked \textbf{M}ixture \textbf{o}f Experts Lo\textbf{RA} (SMoRA), which \textit{treats each rank of LoRA as an expert}. By incorporating an MoE structure within a single LoRA, SMoRA enables dynamic rank-wise sparse activation, as shown in Fig.\ref{fig:compare}c). This allows the model to dynamically and flexibly combine ranks, addressing the rigid knowledge isolation in MoE-based LoRA and enabling fine-grained task adaptation and knowledge sharing.   
However, the dynamic activation mechanism in SMoRA necessitates careful management of rank allocation and computational efficiency to fully harness its potential.
To address these challenges, we adopt a loss-free load-balancing strategy~\cite{wang2024auxiliary}, incorporating an extra expert bias term to ensure balanced rank utilization without modifying the base model. Additionally, we develop a custom CUDA kernel \textit{indexed\_matmul} using TVM~\cite{chen2018tvm}, which performs dynamic matrix multiplications based on router-selected indices, preventing additional computational burden.

Experimental results demonstrate the superior efficiency and performance of SMoRA. Specifically, SMoRA, activating only 8 out of 64 ranks, outperforms a 64-rank LoRA by 1.73\% and an 8-rank LoRA by 11.16\%, highlighting its superior utilization of limited parameters through dynamic rank-wise activation. Additionally, SMoRA surpasses LoRA MoE, which selects the top-1 expert from 8-rank LoRA groups, by 6.13\%, demonstrating its effectiveness in balancing task-specific adaptation and knowledge sharing. 
Our main contribution could be summarized as follows:
\vspace{-1.5mm}
\begin{itemize}[leftmargin=*]
\setlength{\parskip}{3pt}
    \item \textbf{Unified LoRA Framework and Analysis of MoE Granularity Impact}: We establish a unified framework that connects multi-LoRA MoE with standard single-LoRA. Building on this, our in-depth analysis reveals how expert granularity within LoRA-MoE architectures critically impacts model performance and influences the learned parameters' representational capability.
    \item \textbf{Dynamic Rank-wise Activation with SMoRA}: We introduce SMoRA, which integrates MoE into LoRA by treating each rank as an independent expert and dynamically activating only the relevant ranks. This approach effectively balances task-specific adaptation with efficient knowledge sharing, improving performance while maintaining computational efficiency.
    \item \textbf{Superior Performance with Reduced Active Ranks}: Our experiments demonstrate the effectiveness of SMoRA, achieving performance comparable to a full-rank (64-rank) LoRA using only 8 dynamically activated ranks, highlighting its efficiency and effectiveness.
\end{itemize}

\section{Preliminary}
\subsection{Low-Rank Adaptation \& LoRA MoE}
Fine-tuning LLMs with all parameters can be computationally prohibitive, particularly in low-resource scenarios. To mitigate this challenge, \citet{hu2021lora} introduced LoRA, a method that enables efficient fine-tuning by incorporating a small number of trainable parameters. Instead of directly updating the pre-trained weights $\mW_0 \in \mathbb{R}^{d \times d}$, LoRA represents the update $\Delta \mW$ using a low-rank factorization: $\Delta \mW = \mB \mA$, where $\mB \in \mathbb{R}^{d \times r}$, $\mA \in \mathbb{R}^{r \times d}$, and $r \ll d$. The forward computation then becomes:
\begin{equation}
    \vy = \mW_0 \vx + \Delta \mW \vx = \mW_0 \vx + \mB \mA \vx,
\end{equation}
where $\vx \in \mathbb{R}^d$ is the input, and $\vy \in \mathbb{R}^d$ is the output. This approach drastically reduces the number of trainable parameters and the associated computational cost while preserving the model's effectiveness.

Extending this idea, \textbf{LoRA MoE}~\cite{dou2023loramoe,mao2025survey,liu2023moelora,shen2023mixture,zhu2023sira} incorporates multiple LoRA modules (referred to as "experts") and employs an input-dependent router to select or combine these experts dynamically. Formally, the forward pass of LoRA MoE becomes:
\begin{equation}
    \vy = \mW_0 \vx + \sum_{i=1}^n \vg_i(\vx) \mB_i \mA_i \vx.
\end{equation}
where $\mB_i \in \mathbb{R}^{d \times r_i}$, $\mA_i \in \mathbb{R}^{r_i \times d}$, and $\vg_i(\vx) \in [0, 1]$ is the gating function, which controls the activation of the $i$-th expert based on the input $\vx$. The gating pattern can be adjusted to select either sparse activation or soft activation, depending on the specific needs of the model.



\subsection{The Equivalence Between Multi-LoRA MoE and Single-Rank Blockwise Activation}
\label{sec:equivalence}
In this section, we construct a unified framework to establish the equivalence between the multi-LoRA MoE and a single LoRA with block-wise activations.


\begin{property}
    \label{thm:MoE_equivalence}
Let $\mW_0 \in \mathbb{R}^{d \times d}$ denote a base weight matrix, and consider $n$ LoRA experts parameterized by $\{(\mA_i, \mB_i)\}_{i=1}^n$, where $\mA_i \in \mathbb{R}^{d \times r_i}$ and $\mB_i \in \mathbb{R}^{r_i \times d}$. Assume the presence of a gating function $\vg(\vx)$ such that, for each input $\vx \in \mathbb{R}^d$, the vector $\vg(\vx) = (\vg_1(\vx), \vg_2(\vx), \ldots, \vg_n(\vx))^\top$ specifies an activation pattern across the experts. Each $\vg_i(\vx)$ may represent either a sparse activation (e.g., via top-k selection) or a soft activation, with $\vg_i(\vx) \in [0, 1]$ for all $i$. Define:
\[
\tilde{\mA} = \begin{pmatrix} \mA_1 & \mA_2 & \cdots & \mA_n \end{pmatrix} \in \mathbb{R}^{d \times R},
\]
\[
\tilde{\mB} = \begin{pmatrix} \mB_1 \\ \mB_2 \\ \vdots \\ \mB_n \end{pmatrix} \in \mathbb{R}^{R \times d},
\]
where $R = \sum_{i=1}^n r_i$, and let
\[
\mG(\vx) = \mathrm{diag}(\vg_1(\vx) \mI_{r_1}, \; \vg_2(\vx) \mI_{r_2}, \; \ldots, \; \vg_n(\vx) \mI_{r_n}),
\]
where $\mI_{r_i} \in \mathbb{R}^{r_i \times r_i}$ is the identity matrix. The multiplication $\vg_i(\vx) \mI_{r_i}$ expands the scalar $\vg_i(\vx)$ to an $r_i \times r_i$ diagonal matrix, ensuring that the gating weight aligns with the dimensions of the corresponding expert's subspace. Then, \textbf{the output of the multi-LoRA MoE architecture is equivalent to single LoRA with block-wise activation} as:
\begin{equation}
\label{eq:equivalence}
\mW_0 \vx + \sum_{i=1}^n \vg_i(\vx) \mB_i \mA_i \vx = \mW_0 \vx + \tilde{\mB} \mG(\vx) \tilde{\mA} \vx.
\end{equation}
\end{property}
The gating vector in Eq.\ref{eq:equivalence} controls expert activation through its diagonal matrix form. This enables the term to function as a block-activated LoRA module, where each expert corresponds to a submatrix block, and its activation is governed by the gating mechanism. In essence, the MoE architecture functions as a single LoRA module with sparse, input-dependent block-wise activation.

\subsection{Motivation}
\label{subsec:motivation}
Property~\ref{thm:MoE_equivalence} establishes that constructing multiple LoRA experts is equivalent to a single LoRA framework with block-wise activation, where each expert corresponds to a partition of the rank parameter. This equivalence naturally leads us to question how the granularity of this rank partitioning impacts performance, especially on diverse downstream tasks.

To investigate this, we conducted experiments on the heterogeneous FLAN-v2 dataset~\cite{longpre2023flan}. We initialized LoRA with a total rank of 64 and compared different partitioning schemes while keeping the number of activated ranks constant at 16. As shown in Fig.\ref{fig:motivation} a), our results demonstrate a clear trend: finer partitioning granularity leads to improved performance, even when the total parameter count and the number of activated parameters remain constant.

To understand the underlying structural changes driving this performance gain, in Fig.\ref{fig:motivation} b), we visualized the rank-wise parameter similarity using $[A,B^T][A,B^T]^T$ heatmaps of a layer in LLM. 
These visualizations reveal distinct structural patterns. Coarse-grained partitioning (e.g., 16 ranks per expert in a 16x4 configuration) results in prominent block-diagonal structures, indicating that the learned parameters within each large expert block are highly similar and thus more concentrated. This internal homogeneity, while ensuring separation between expert blocks, limits the diversity of knowledge each coarse expert can capture, thereby impeding the model's flexibility in learning varied information.
Instead, with fine-grained partitioning (such as each rank-1 component acting as an expert in a 1x64 setup), a prominent main diagonal emerges. This, coupled with increasingly diffuse and complex off-diagonal correlations, points to a substantial reduction in the learned correlation between parameters of different fine-grained experts. These experts thus maintain their distinctiveness rather than forming larger correlated groups, a characteristic that enables the model to capture a wider and more flexible array of knowledge by leveraging more specialized, less redundant expert contributions.
Furthermore, this fine-grained structure enables more precise control over knowledge selection and sharing via the routing mechanism and allows for a greater variety of expert combinations during inference. 
Therefore, these findings motivate our approach: we embed the MoE structure directly within LoRA. Since the parameters of each rank in LoRA represent its smallest constituent unit, we naturally treat each rank as an individual expert. This rank-wise activation achieves the finest possible partitioning granularity, offering a principled way to balance task specialization and knowledge sharing.

\section{Methodology}

\subsection{Architecture of SMoRA}
\label{sec:framework}

\begin{figure}
    \centering
\includegraphics[width=\linewidth]{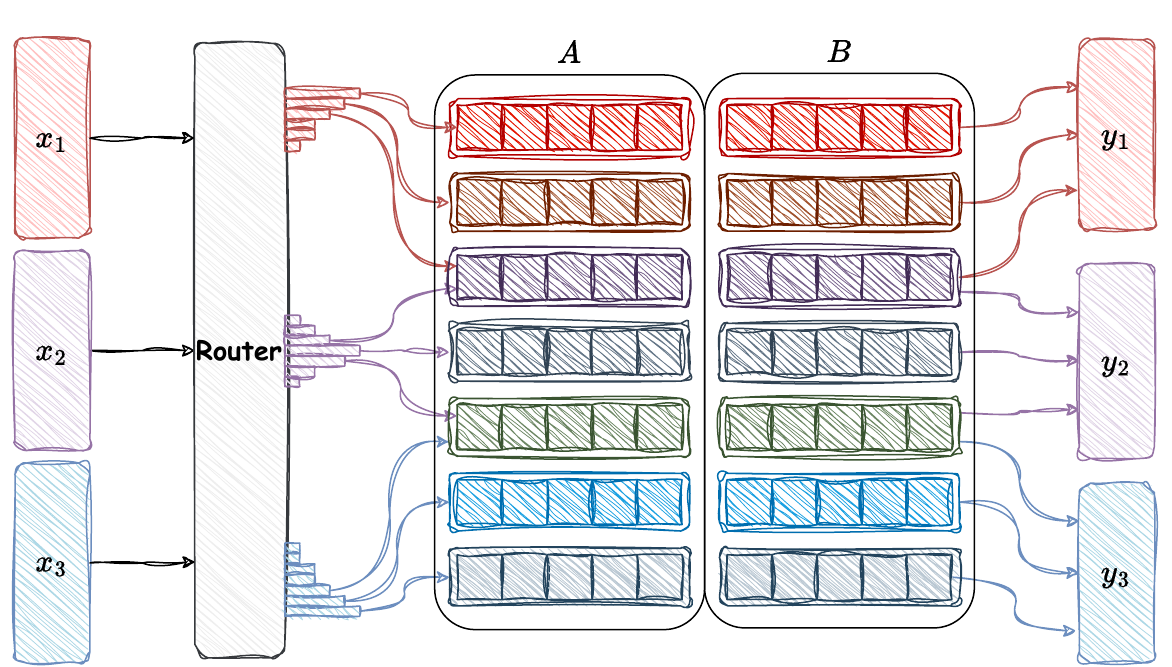}
    \caption{\textbf{Overall Framework of SMoRA.} SMoRA builds upon Vanilla LoRA by adding a router that selects the appropriate rank for activation based on each token. The framework is outlined in \S\ref{sec:framework}, the load balancing strategy is discussed in \S\ref{sec:load_balance}, and the custom CUDA kernel for efficient sparse matrix computations is described in \S\ref{sec:cuda_operator}.}
    \label{fig:framework}
\end{figure}

The overall architecture of SMoRA is illustrated in Fig.\ref{fig:framework}. Building upon pre-trained weights $\mW_0 \in \mathbb{R}^{d \times d}$, SMoRA introduces additional trainable low-rank matrices $\mB \in \mathbb{R}^{d \times r}$ and $\mA \in \mathbb{R}^{r \times d}$, as in vanilla LoRA, where $r$ denotes the rank. Additionally, SMoRA extends LoRA by treating each rank as an individual ``expert" and incorporating a sparse MoE mechanism. A gating function $\vg(\vx) \in \mathbb{R}^r$ determines the importance of each rank for a given input $\vx$, enabling dynamic sparse activation of the most relevant ranks via \textbf{top-k routing}. The gating matrix $\mG(\vx) = \mathrm{diag}(\vg_1(\vx), \vg_2(\vx), \ldots, \vg_r(\vx))$ is diagonal, where $\vg_i(\vx)$ represents the gating score for the $i$-th rank expert. The forward pass of SMoRA is formulated as:
\begin{equation}
\label{eq:smora_cal}
    \vy =\mW_0 \vx + \mB \mG(\vx) \mA \vx,
\end{equation}
where $\mG(\vx) \in \mathbb{R}^{r \times r}$ selectively activates the top-k ranks based on the gating scores $\vg(\vx)$. Specifically, ranks with the highest rating scores are activated, while the rest are ignored (i.e., their contributions are set to zero). This sparse activation mechanism focuses computation on the most relevant ranks for each token, improving both efficiency and model adaptability.

The gating scores $\vg(\vx)$ are computed as follows:
\begin{equation}
    \label{eq:original_gating}
    \vg(\vx) = \mathrm{Softmax}(\mathrm{TopK}(\mW_g \vx)),
\end{equation}
where $\vx \in \mathbb{R}^d$ is the input, $\mW_g \in \mathbb{R}^{r \times d}$ is a learnable projection matrix, and $\mathrm{TopK}(\cdot)$ retains only the top-k largest values, setting others to \(-\infty\) before applying the Softmax. The sparsity level is controlled by the hyperparameter $k$, which defines the number of active ranks for each token.

By integrating LoRA with fine-grained rank-wise sparse activation, SMoRA dynamically combines parameters of different ranks based on the input, enabling efficient parameter decoupling and knowledge sharing across tasks to achieve superior performance on heterogeneous tasks.

\subsection{Load Balance with Loss-Free Balancing}
\label{sec:load_balance}
To ensure load balancing during expert routing without introducing additional training overhead and to guarantee seamless adapter deployment across different models, we follow \citet{wang2024auxiliary} by incorporating an auxiliary bias term in the gating output. Specifically, the gating function $\vg(\vx)$ in Eq.\ref{eq:original_gating} is modified as:
\begin{equation}
    \label{eq:bias_routing}
    \vg(\vx) = \mathrm{Softmax}(\mathrm{TopK}(\mW_g \vx + \vb)),
\end{equation}
where $\vb\in \mathbb{R}^r$ is a bias term that adjusts the routing probabilities to achieve balanced expert utilization. 

During the training phase, the bias term is updated according to the formula:
\begin{equation}
    b_i = b_i + u*sign(e_i),
\end{equation}
where $e_i = \bar{c}_i - c_i$ represents the difference between the average number of token $\bar{c}_i$ and the number of assigned tokens $c_i$ for expert $i$. The update rate $u$ is a small positive constant that controls the magnitude of the bias adjustment. This loss-free balancing mechanism effectively promotes uniform expert utilization without requiring additional loss terms in the overall training objective. 

This training-free load-balancing approach aligns well with the philosophy of PEFT methods, which emphasizes modularity by avoiding modifications to the base model. By sidestepping the need for model-specific balancing losses, our method preserves the plug-and-play nature of adapters, enabling them to be directly loaded as parameters into the linear layers of the model without altering the underlying architecture or training process of the base model.

\subsection{Implementation of Indexed Sparse LoRA Calculation.}
\label{sec:cuda_operator}

To improve the efficiency of sparse matrix operations in SMoRA, we developed \textit{indexed\_matmul}, a custom CUDA kernel leveraging TVM~\cite{chen2018tvm}. This kernel utilizes top-K indices from the router's gating function to selectively compute only with the active rows and columns of the LoRA matrices. This targeted computation method primarily avoids two common sources of inefficiency: first, the substantial computational costs inherent in dense matrix operations, by processing only essential data; and second, the latency from iterative processing, as it bypasses the explicit for-loops typically employed in many MoE frameworks to handle individual experts. For a detailed implementation and further discussion of TVM optimizations, please see Appendix~\ref{app:tvm}.

\begin{table*}[h!]
\centering
\caption{\textbf{Performance Across Task Clusters in FLAN-v2.} The best performance for each task is highlighted in bold, and the second-best performance is underlined. The performance of our proposed SMoRA is highlighted in gray. The full results of each task are shown in Appendix~\ref{sec:full_res}}
\label{tab:flan}
\resizebox{.85\linewidth}{!}{%
\begin{tabular}{lccccccccc}
\toprule
\textbf{Task}       & \textbf{SMoRA-64-8} & \textbf{LoRA-64} & \textbf{LoRA-8} & \textbf{MoE-Top1} & \textbf{MoE-Top2} & \textbf{MoE-Soft} & \textbf{SMEAR} & \textbf{HydraLoRA} & \textbf{MosLoRA}\\
\midrule
\multicolumn{10}{c}{\textbf{\textit{w/ Llama2-7b}}} \\ \hline
\textbf{Struct to Text Rouge-1} & \textbf{\cellcolor[gray]{0.9}62.7} & \underline{62.4} & 59.2 & 60.2 & 61.2 & 60.8 & 59.6 & 59.9 & 62.1 \\
\textbf{Struct to Text Rouge-2} & \textbf{\cellcolor[gray]{0.9}37.6} & \underline{37.2} & 32.7 & 34.0 & 35.3 & 35.3 & 33.5 & 34.0 & 36.3 \\
\textbf{Struct to Text Rouge-L} & \textbf{\cellcolor[gray]{0.9}55.4} & \underline{55.0} & 52.6 & 54.2 & 54.5 & 54.5 & 53.0 & 52.9 & 54.9 \\
\textbf{Translation BLEU} & \cellcolor[gray]{0.9}13.0 & 13.1 & \underline{13.3} & 12.6 & 12.8 & 13.1 & \underline{13.3} & \underline{13.3} & \textbf{13.4} \\
\midrule
\textbf{COMMONSENSE} & \textbf{\cellcolor[gray]{0.9}67.0} & \underline{66.0} & 61.5 & 60.5 & 64.0 & 62.0 & 61.0 & 63.0 & 64.5 \\
\textbf{SENTIMENT} & \textbf{\cellcolor[gray]{0.9}91.0} & 90.0 & \underline{90.5} & 90.0 & \underline{90.5} & \underline{90.5} & \underline{90.5} & \underline{90.5} & \textbf{91.0} \\
\textbf{READING Comp.} & \textbf{\cellcolor[gray]{0.9}52.7} & 48.7 & 44.3 & 48.0 & \underline{50.3} & 49.0 & 50.0 & 46.7 & 48.0 \\
\textbf{CLOSE-BOOK QA} & \underline{\cellcolor[gray]{0.9}55.5} & 54.5 & 49.5 & 51.0 & \textbf{56.5} & 53.0 & 51.0 & 51.5 & 55.0 \\
\textbf{COREFERENCE} & \cellcolor[gray]{0.9}59.0 & 61.0 & 56.0 & 57.0 & 52.0 & 57.0 & 59.0 & \textbf{62.0} & \underline{61.0} \\
\textbf{READ. COOMP. W/ COM} & \textbf{\cellcolor[gray]{0.9}71.0} & \underline{67.0} & 53.0 & 54.0 & 63.0 & 60.0 & 61.0 & 54.0 & 66.0 \\
\textbf{PARAPHRASE} & \underline{\cellcolor[gray]{0.9}68.5} & \textbf{70.0} & 56.0 & 60.5 & 65.0 & 62.5 & 62.5 & 60.5 & 66.5 \\
\textbf{NLI} & \underline{\cellcolor[gray]{0.9}71.8} & 69.2 & 65.7 & 57.4 & 68.1 & 67.5 & 67.7 & 68.3 & \textbf{72.6} \\
\midrule
\textbf{AVERAGE} & \textbf{\cellcolor[gray]{0.9}58.8} & \underline{57.8} & 52.9 & 55.4 & 56.1 & 53.3 & 55.2 & 54.7 & 57.6 \\
\midrule
\multicolumn{10}{c}{\textbf{\textit{w/ Llama2-13b}}} \\ \hline
\textbf{Struct to Text Rouge-1} & \underline{\cellcolor[gray]{0.9}63.5} & 63.4 & 60.1 & 62.2 & 61.2 & 62.8 & 62.3 & 61.1 & \textbf{64.0} \\
\textbf{Struct to Text Rouge-2} & \cellcolor[gray]{0.9}38.5 & \underline{38.7} & 34.1 & 36.4 & 35.4 & 37.3 & 35.9 & 35.1 & \textbf{39.3} \\
\textbf{Struct to Text Rouge-L} & \underline{\cellcolor[gray]{0.9}55.7} & 55.5 & 53.6 & \underline{55.7} & 55.5 & 55.6 & 55.5 & 54.8 & \textbf{56.7} \\
\textbf{Translation BLEU} & \cellcolor[gray]{0.9}14.3 & \underline{14.6} & \underline{14.6} & 14.1 & 13.8 & 14.5 & \textbf{14.7} & \underline{14.6} & 14.2 \\
\midrule
\textbf{COMMONSENSE} & \textbf{\cellcolor[gray]{0.9}70.0} & 68.5 & \underline{69.0} & 67.0 & 68.5 & 67.5 & \underline{69.0} & \underline{69.0} & 67.0 \\
\textbf{SENTIMENT} & \cellcolor[gray]{0.9}91.0 & 90.5 & \textbf{92.5} & \underline{92.0} & 91.5 & 91.5 & 90.5 & \textbf{92.5} & 90.5 \\
\textbf{READING Comp.} & \textbf{\cellcolor[gray]{0.9}55.3} & 53.7 & 51.3 & 52.0 & 50.7 & 51.7 & 52.0 & 51.7 & \underline{54.0} \\
\textbf{CLOSE-BOOK QA} & \textbf{\cellcolor[gray]{0.9}61.5} & 59.5 & 57.0 & 60.0 & 58.0 & \underline{60.5} & 59.0 & 58.0 & 59.5 \\
\textbf{COREFERENCE} & \textbf{\cellcolor[gray]{0.9}68.0} & \underline{67.0} & 62.0 & 62.0 & 66.0 & 64.0 & 60.0 & 63.0 & 64.0 \\
\textbf{READ. COOMP. W/ COM} & \underline{\cellcolor[gray]{0.9}73.0} & 71.0 & 66.0 & 66.0 & 63.0 & 69.0 & 67.0 & 68.0 & \textbf{74.0} \\
\textbf{PARAPHRASE} & \textbf{\cellcolor[gray]{0.9}67.5} & \underline{67.0} & 64.5 & 65.5 & 66.5 & 66.5 & \underline{67.0} & \underline{67.0} & \underline{67.0} \\
\textbf{NLI} & \underline{\cellcolor[gray]{0.9}75.1} & 74.7 & 70.7 & 74.2 & 73.3 & 75.0 & \textbf{75.7} & 73.1 & 74.0 \\
\midrule
\textbf{\textbf{AVERAGE}} & \textbf{\cellcolor[gray]{0.9}61.1} & 60.3 & 58.0 & 58.9 & 58.6 & 59.7 & 59.1 & 59.0 & \underline{60.4} \\

\bottomrule
\end{tabular}%
}
    \vspace{-4mm}
\end{table*}

\begin{table}[h!]
\centering
\caption{\textbf{Accuracy Across Multi-Domain Benchmark.} The best performance for each domain is highlighted in bold, and the second-best performance is underlined. The performance of our proposed SMoRA is highlighted in gray.}
\label{tab:multidomain}
\resizebox{\linewidth}{!}{%
\begin{tabular}{lcccccccc}
\toprule
\multirow{2}{*}{\textbf{Method}} & \multirow{2}{*}{\textbf{GSM8K}} & \multirow{2}{*}{\textbf{MMLU}} & \multirow{2}{*}{\textbf{Law}} & \multirow{2}{*}{\textbf{Medical}} & \multicolumn{2}{c}{\textbf{HumanEval}} & \multirow{2}{*}{\textbf{Average}} \\
\cline{6-7}
& & & & & \textbf{Pass@1} & \textbf{Pass@10} & \\
\midrule
\textbf{LoRA-64 }        & \textit{34.80} & 41.45 & 41.79 & \textbf{44.61} &\textbf{18.90} & 29.88 & 35.24 \\
\textbf{MoE-Top1}        & 26.46 & 38.53 & 38.41 & 34.35 & 15.85 & 30.49 & 30.68 \\
\textbf{MoE-Top2}       & 31.39 & 39.92 & 39.65 & 39.71 & 15.24 & 25.00 & 31.82 \\
\textbf{MoE-Soft}        & 28.89 & 41.52 & 43.98 & 41.57 & 15.85 & \underline{30.97} & \underline{33.79} \\
\textbf{SMEAR}           & 27.90 & 40.68 & \textbf{45.35} & 38.31 & 16.46 & 26.83 & 32.59 \\
\textbf{HydraLoRA}       & 27.14 & 41.54 & 42.38 & 39.50 & 15.85 & 28.05 & 32.41 \\
\textbf{MoSLoRA}         & 33.66 & 40.65 & 43.32 & 41.38 & \underline{18.29} & 30.49 & 34.63 \\
\midrule
\textbf{SMoRA}          & \cellcolor[gray]{0.9}\textbf{34.95} & \cellcolor[gray]{0.9}\textbf{41.70} & \cellcolor[gray]{0.9}\underline{45.20} & \cellcolor[gray]{0.9}\underline{43.81} & \cellcolor[gray]{0.9}\textbf{18.90} & \cellcolor[gray]{0.9}\textbf{35.37} & \cellcolor[gray]{0.9}\textbf{36.66} \\
\bottomrule
\end{tabular}%
}
\vspace{-4mm}
\end{table}

\section{Experiments}
\paragraph{Experiment Setup.}
To validate the efficacy of current PEFT methods in multi-task learning scenarios, we leverage Llama-2-\{7b,13b\}~\cite{touvron2023llama} as the base models and train a range of LoRAs for a spectrum of tasks. We select a portion of the \textbf{Flan-v2 datasets}, which contains 48 tasks covering Natural Language Understanding (NLU) and Natural Language Generation (NLG), that can be grouped into 10 distinct task clusters. 
To further evaluate the effectiveness of our proposed method in domain-specific tasks, we constructed a \textbf{multi-domain benchmark} and tested it using Llama-2-7b. This benchmark spans general language, medical, legal, mathematical, and code generation tasks, following the approach outlined in \cite{tian2024hydralora}. The evaluation set includes: (1) MMLU~\cite{chen2021evaluating} for general ability, (2) Law tasks from MMLU, (3) Medical tasks from MMLU, (4) the GSM8K~\cite{cobbe2021training} evaluation set for mathematics, and (5) HumanEval~\cite{chen2021evaluating} for code generation.
Details of the tasks can be found in Appendix~\ref{app:eval_bench}.
Additionally, we evaluate the performance of the Llama3-Instruct model~\cite{dubey2024llama} after training on code-related datasets, specifically assessing its performance on multiple data analysis tasks using Python libraries. The results of these evaluations are provided in Appendix~\ref{app:python_results}.


\paragraph{Baseline Methods.}
We compare the proposed method against several state-of-the-art PEFT methods, including (1) LoRA with varying numbers of ranks; (2) LoRA MoE with top-1 and top-2 routing, where each expert is a LoRA with rank 8; (3) LoRA MoE with soft routing, which computes a weighted average of the experts' outputs based on the gating scores; (4) SMEAR, which performs LoRA fusion based on the gating scores; (5) HydraLoRA, which leverages an asymmetric structure consisting of one $\mA$ matrix and four $\mB$ matrices, each with rank 8; (6) MoSLoRA, which incorporates a mixture matrix between the LoRA matrices $\mA$ and $\mB$. Detailed implementations of each baseline method can be found in Appendix~\ref{app:baseline_methods}.

\paragraph{Implementation Details of SMoRA.}
The total rank of all LoRA MoE baselines is set to 64. SMoRA follows this setting with a total rank of 64, activating 8 rank parameters with an additional router. The update rate $u$ of the expert routing bias in Eq.\ref{eq:bias_routing} is set to $1 \times 10^{-5}$ for all SMoRA experiments. Additionally, the initialization of SMoRA's $\mA$ and $\mB$ matrices is identical to that of vanilla LoRA, ensuring consistency in the base parameter setup. An ablation study on the effect of total activation rank in downstream tasks is provided in Appendix~\ref{app:ablation_rank}.

\begin{figure*}
    \centering
\includegraphics[width=.95\linewidth]{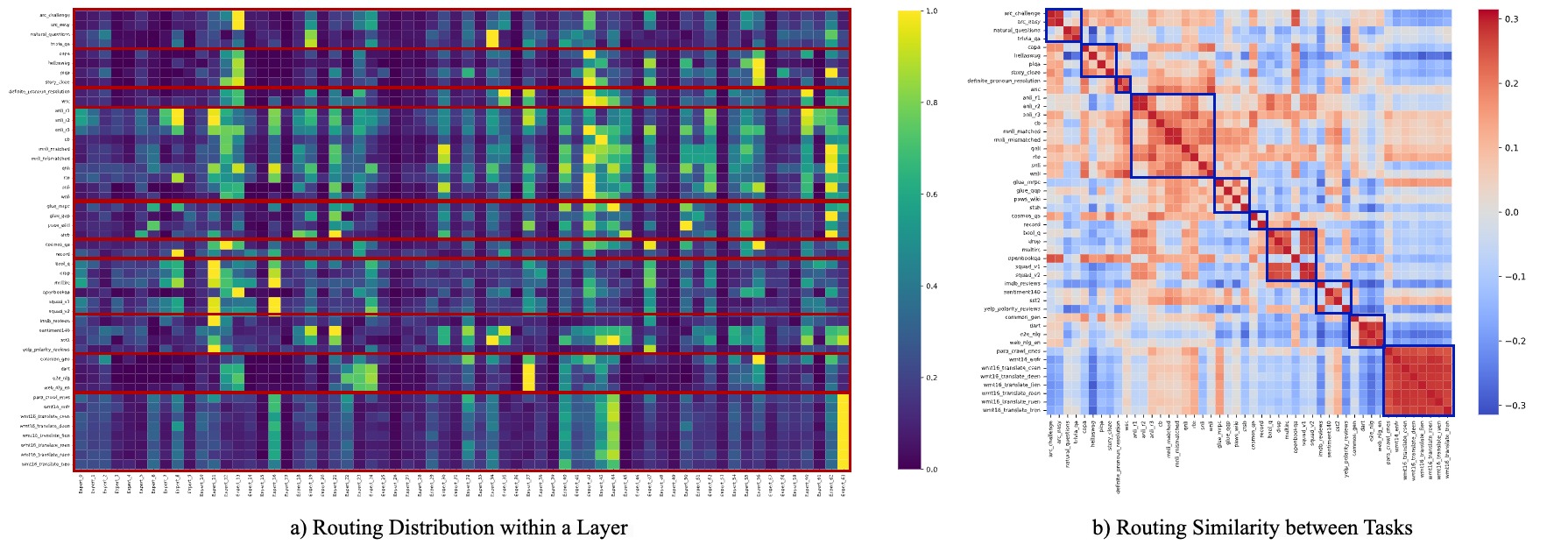}
    \vspace{-4mm}
    \caption{\textbf{Visualization of Routing Distribution.} Tasks from the same domain are grouped in square brackets. The X-axis in Figure a) denotes the expert index.}
    \label{fig:routing}
        \vspace{-4mm}
\end{figure*}

\begin{figure}
    \centering
\includegraphics[width=\linewidth]{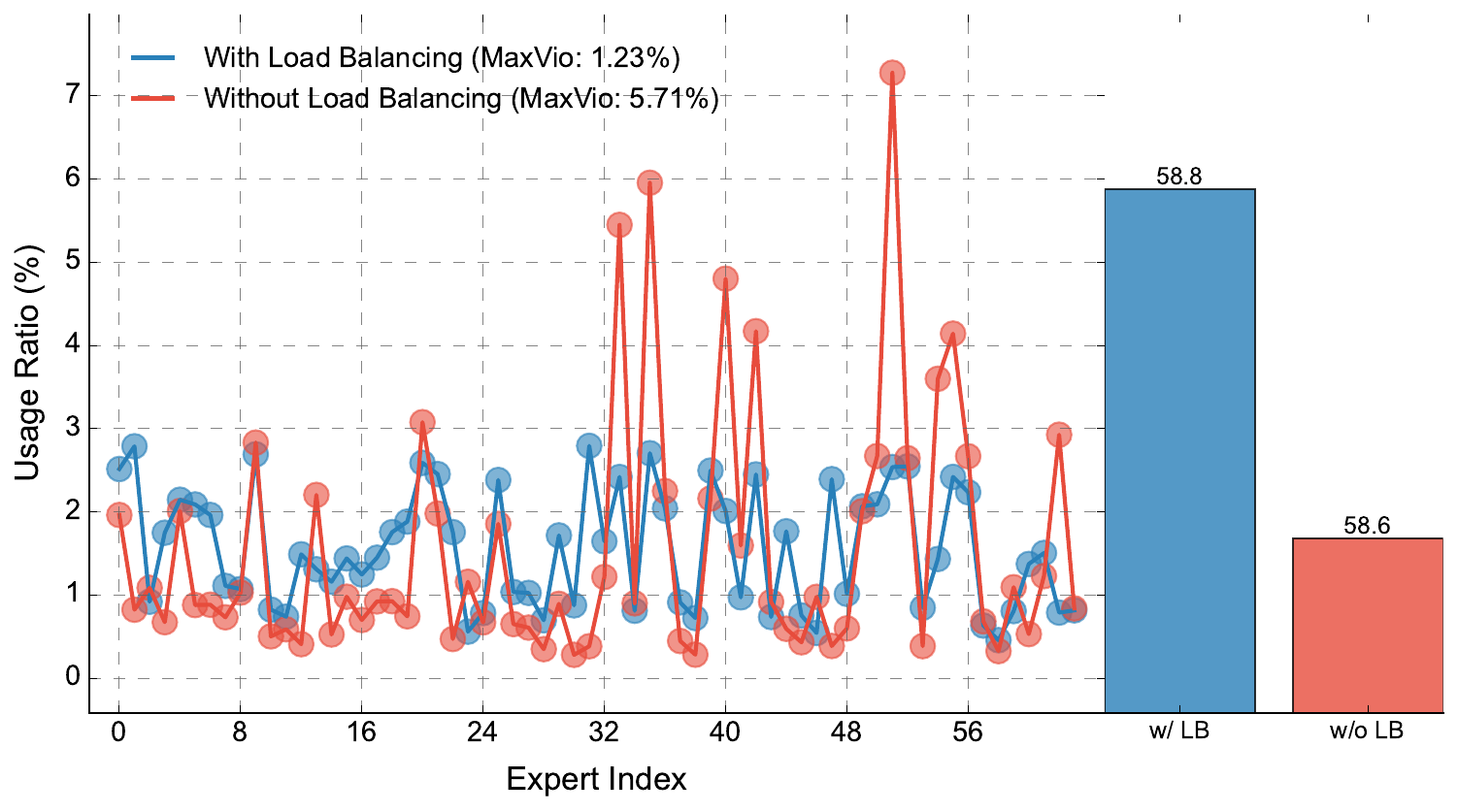}
        \vspace{-4mm}
    \caption{\textbf{The Impact of Load Balancing Strategy on Expert Load Distribution.} The left figure compares expert load distribution with and without a load balancing strategy, while the right shows the resulting performance differences.}
        \label{fig:load}
    \vspace{-6mm}
\end{figure}

\paragraph{Main Results on Flan-v2}
The main results for Flan-v2 are presented in Tab.\ref{tab:flan}. Based on these results, we make the following observations:
(1) The performance of Vanilla LoRA improves gradually as the rank increases. However, due to task interference, it performs worse than the MoE and SMoRA methods when compared under the same number of activated parameters.
(2) Due to the partitioning of parameters, MoE methods perform better than LoRA counterparts with the same number of activated parameters. However, as they fail to fully utilize shared knowledge across different tasks, their performance remains suboptimal.
(3) By employing rank-wise activation and fine-grained parameter partitioning, SMoRA effectively decouples parameters, striking a balance between task interference and knowledge sharing. This allows SMoRA to achieve the best performance. 
(4) Notably, SMoRA, which activates only 8 out of 64 ranks in LoRA, outperforms the full Vanilla LoRA model (with all 64 ranks activated), achieving a performance improvement of 1.73\% in the Llama2-7b experiment and 1.33\% in Llama2-13b. Furthermore, compared to a LoRA model with 8 activated ranks, SMoRA demonstrates an 11.16\% improvement in Llama2-7b and 5.34\% in Llama2-13b. Additionally, SMoRA outperforms LoRA MoE, which treats every 8 ranks as a separate expert and activates the top-1 expert, achieving a 6.13\% improvement in Llama2-7b and 3.74\% in Llama2-13b. These results emphasize SMoRA's superior effectiveness, even with a smaller subset of activated ranks.
We also compared the activated parameter count across different methods in Appendix~\ref{app:para_count}.

\paragraph{Results on Multi-Domain Benchmark.}
Tab.\ref{tab:multidomain} presents the results of different methods on the multi-domain benchmark. From the results, we make the following observations:
(1) When LoRA is fully activated, it performs well on tasks like mathematics and code generation, benefiting from the increased number of activated parameters. However, its performance drops significantly on tasks such as MMLU, LAW, and Medical, due to task interference.
(2) The MoE-based LoRA method struggles with domain-specific multi-domain tasks because it lacks the flexibility to effectively combine parameters and knowledge.
(3) In contrast, SMoRA excels by leveraging fine-grained parameter decoupling, which enables flexible combinations of parameters and knowledge. This results in superior performance across all tasks, especially evident in MMLU, LAW, Medical, and HumanEval Pass@10, underscoring the effectiveness of our approach.

\paragraph{Routing Visualization.}  
To better understand the routing distribution of SMoRA, we visualize it by selecting a model layer and plotting the routing patterns for a portion of the evaluation set, categorized by each task. 
From Fig.\ref{fig:routing}a), we observe that rank-wise routing effectively decouples parameters, with each task displaying a distinct expert distribution. This indicates that different tasks can activate unique rank parameters, thereby minimizing task interference. Furthermore, tasks with similarities exhibit comparable routing distributions. Fig.\ref{fig:routing}b) presents heatmaps of the routing distributions across various tasks, revealing that tasks within the same cluster share similar routing patterns. This observation suggests that SMoRA enhances the sharing of related knowledge. Overall, these results demonstrate SMoRA’s strength in fine-grained knowledge sharing and parameter decoupling.

\paragraph{Ablation on Load Balance.}
In Fig.\ref{fig:load}, we select a specific layer and illustrate the load distribution across all ranks, both with and without the auxiliary load balancing strategy. When the load balancing strategy is applied, the load is evenly distributed across ranks, indicating a relatively balanced workload. In contrast, without the load-balancing strategy, certain ranks experience significantly higher loads, leading to an imbalanced workload distribution. To quantitatively evaluate the load balance, we calculate the MaxVio (maximal violation) value~\cite{wang2024auxiliary}, defined as the difference between the maximum load and the average load, with smaller values indicating a more balanced distribution. With the load balancing strategy, the MaxVio is significantly reduced to 1.23, compared to 5.71 without the strategy. This stark contrast further highlights the effectiveness of the adopted load-balancing strategy in achieving a more uniform load distribution. As shown in the right part of Fig.\ref{fig:load}, applying the load-balancing strategy results in a 0.34\% performance improvement. In Appendix~\ref{app:routing_ablation}, we further visualize the routing distribution under the ablation of the load-balancing strategy. The results clearly demonstrate that with the load-balancing strategy removed, the routing becomes unevenly distributed, heavily favoring a subset of experts. This imbalance highlights the critical role of the load-balancing mechanism in ensuring a fair and efficient allocation of tasks across all available experts.

\begin{figure}
    \centering
\includegraphics[width=.9\linewidth]{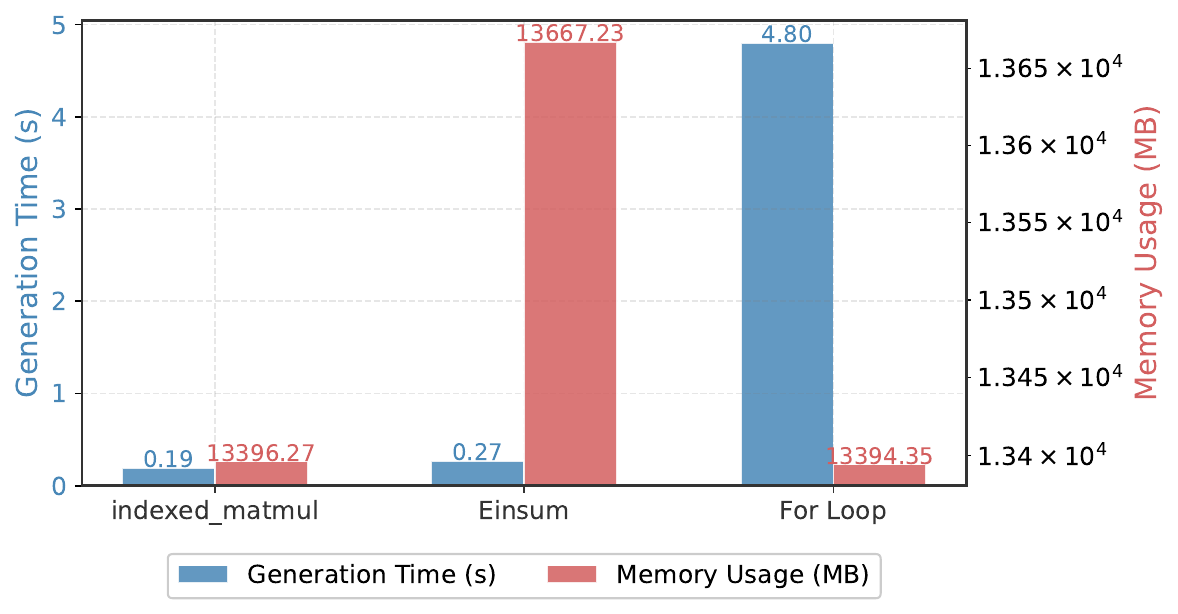}
    \caption{\textbf{Average Time and Average Peak GPU Memory Usage for Generating One Token.} We report results based on a subset of the evaluation set, showing the average time taken to generate the first token (in seconds) and the average peak GPU memory usage (in MB) for different methods.}
    \label{fig:mem_time}
        \vspace{-4mm}
\end{figure}

\paragraph{Effectiveness of the Implemented \textit{indexed\_matmul}.}
In our approach, the router selects the appropriate ranks to participate in the computation for each token. As a result, efficiently implementing sparse matrix operations is beyond the capabilities of current PyTorch operators. Typically, previous LoRA-MoE implementations iterate over each expert and construct the corresponding inputs in a ``\textit{for loop}" style. However, in SMoRA, the large number of experts makes direct for-loop iteration prohibitively time-consuming. Another approach involves using \textit{einsum} to construct the intermediate LoRA matrices for each token, but this consumes significant GPU memory, making it infeasible for large-scale training and inference. To overcome this, we developed a custom CUDA kernel, \textit{indexed\_matmul}, implemented using TVM, to accelerate sparse matrix computations. Specifically, given two matrices $\vx$ and $\mA$, we use the index to select the corresponding rows in $\mA$ to multiply with $\vx$ (while similarly selecting the corresponding columns in $\mB$ for computation). This approach enables efficient and flexible sparse matrix multiplication. We evaluated 100 data points to measure the time taken to generate the first token during inference and analyzed the peak GPU memory usage across various methods. As shown in Fig.\ref{fig:mem_time}, the \textit{indexed\_matmul} method strikes a balance between computational efficiency and memory overhead, outperforming other implementations.

\section{Related Work}
\label{sec:related_work}
PEFT methods~\cite{hu2021lora,zhang2023adalora,liu2022few,ding2023sparse} with their low-rank decomposition have substantially reduced the computational cost of adapting LLMs. However, these techniques generally yield static parameters after training, which limits their ability to dynamically adjust to diverse inputs, a crucial feature for heterogeneous downstream tasks.

LoRA-MoE approaches~\cite{liu2023moelora,dou2023loramoe,muqeeth2023soft,li2024mixlora,tian2024hydralora} typically use routing mechanisms to select or combine different LoRA modules, treating them as experts. However, the coarse-grained expert division can restrict fine-grained, input-specific adjustments, thereby potentially hindering adaptability and optimal performance on diverse tasks. For a more comprehensive discussion, please refer to Appendix~\ref{app:related_work}.

\section{Conclusion}
This paper demonstrated that multi-LoRA MoE can be unified with single LoRA through block-wise activation. Our empirical analysis further revealed that finer-grained rank activation enhances performance on heterogeneous tasks and expands parameter representational capacity. Based on these insights, we introduced SMoRA, which embeds MoE within LoRA for fine-grained, rank-wise activation. Experimental results show that SMoRA achieves superior performance in multi-task learning with fewer activated parameters.


\section*{Limitations}
One potential limitation of SMoRA's fine-grained expert partitioning, despite its theoretical FLOPs reduction, pertains to its practical execution efficiency on highly parallel hardware like GPUs. The fine-grained approach leads to highly dispersed matrix computations. While theoretically requiring fewer operations, this fragmentation can result in underutilization of the large compute cores prevalent in modern GPUs, potentially yielding practical speedups that are less pronounced than what theoretical FLOPs might suggest. Conversely, this characteristic could be advantageous for edge devices. The typically smaller compute cores in edge hardware might align better with SMoRA's fine-grained computational tasks, potentially offering more significant relative efficiency gains in such scenarios.


\bibliography{custom}

\newpage
\appendix
\onecolumn

\section{Ablation on Activated Rank Number}
\label{app:ablation_rank}
\begin{figure}
    \centering
\includegraphics[width=.75\linewidth]{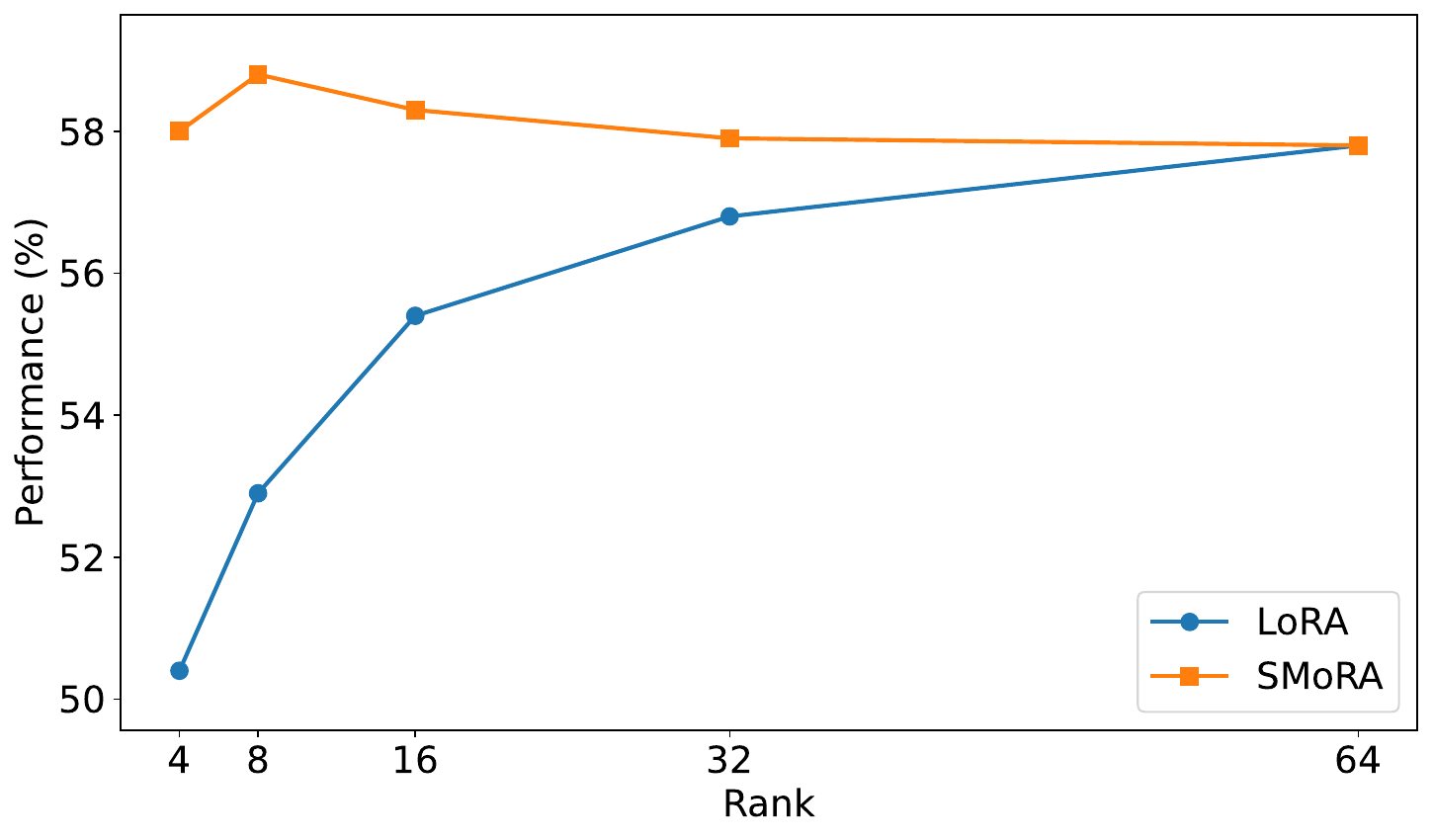}
        \vspace{-4mm}
    \caption{\textbf{Ablation on Activate Rank for Vanilla LoRA and SMoRA.}}
    \label{fig:rank_ablation}
        \vspace{-4mm}
\end{figure}

In Fig.\ref{fig:rank_ablation}, we compare the performance of Vanilla LoRA and SMoRA under different numbers of activated ranks. From the figure, we have the following observations: 
(1) The performance of LoRA improves as the number of total ranks increases; 
2) For SMoRA, the best performance is achieved by activating 8 ranks, as the number of activated ranks governs the granularity of information sharing between tasks. In our setup, which consists of 48 tasks divided into 10 task clusters, the differences among tasks can be substantial. Activating too many ranks may force excessive knowledge sharing, potentially leading to task interference. Conversely, using too few shared ranks could result in insufficient utilization of knowledge. Therefore, selecting 8 ranks for activation strikes the optimal balance, effectively managing the trade-off between inter-task information sharing and interference. 
(3) When SMoRA activates all parameters, it degenerates into Vanilla LoRA, resulting in identical performance. 
(4) With the same number of activated ranks, SMoRA consistently outperforms LoRA.

\section{Detailed Related Work}
\label{app:related_work}
\subsection{PEFT Methods}
Fine-tuning large language models (LLMs) requires substantial computational resources, making it impractical in many scenarios. To address this, PEFT methods update only a small subset of parameters while keeping most of the model frozen. Among these, LoRA~\cite{hu2021lora} is widely used, reducing trainable parameters by decomposing weight updates into the product of two low-rank matrices, improving efficiency without sacrificing performance. Several extensions of LoRA, such as AdaLoRA~\cite{zhang2023adalora}, IA3~\cite{liu2022few}, and SoRA~\cite{ding2023sparse}, aim to enhance its flexibility, dynamic rank allocation, and efficiency. However, these methods fix parameters after training, preventing dynamic selection based on input, which limits adaptability in downstream heterogeneous tasks.

\subsection{LoRA with MoE}
Some works apply MoE-based methods specifically for multi-task or domain-specific learning. For example, MoELoRA~\cite{liu2023moelora} focuses on multi-task learning with explicit task labels in medical applications, while LoRA-MoE~\cite{dou2023loramoe} with soft routing introduces gating mechanisms to allocate LoRA experts during fine-tuning, alleviating catastrophic forgetting. Similarly, SMEAR~\cite{muqeeth2023soft} achieves a dynamic fusion of LoRA experts, allowing the model to adjust its parameters adaptively during different tasks. MixLoRA~\cite{li2024mixlora} employs a top-k routing strategy, typically selecting the top-2 LoRA experts for multi-task learning, while HydraLoRA~\cite{tian2024hydralora} proposes an asymmetric structure for its MoE design, further improving parameter utilization in multi-task scenarios. However, these methods rely on coarse-grained expert partitioning, limiting the flexibility of parameter combinations and hindering the model's ability to dynamically adjust its internal representations. This coarse granularity leads to suboptimal performance, as it fails to fully exploit the potential for more precise and adaptive knowledge sharing across tasks.

In addition to training-time MoE, several post-training methods have also been explored~\cite{zhao2024retrieval,wu2023mole,xu2024meteora}. Techniques such as MoLE propose first training individual LoRA modules for each specific task and subsequently training an additional router to control task-specific activation during inference. However, these post-training MoE designs are largely focused on inference-time flexibility and are less relevant to training-time adaptive methods, which are the main focus of this paper.

\section{Evaluation Benchmark Details}
\label{app:eval_bench}
In our experiments, we evaluate the proposed SMoRA across three benchmarks:
(1) FLAN-v2, which assesses the natural language understanding (NLU) and natural language generation (NLG) capabilities of large language models (LLMs);
(2) Multi-Domain Benchmark, which evaluates the model's performance across multiple vertical domains, including math, general reasoning, law, medicine, and code generation;
(3) The DS-1000 dataset, which is specifically designed to test the model's abilities in code generation.
Details of these benchmarks are provided in Table~\ref{tab:benchmark_summary}. Below, we provide a detailed description of each dataset and discuss its relevance to our evaluation.

\begin{table}[htbp]
\centering
\caption{Summary of Evaluation Benchmarks}
\label{tab:benchmark_summary}
\resizebox{\textwidth}{!}{%
\begin{tabular}{lccc}
\midrule
\textbf{Aspect} & \textbf{Muti-task Language Benchmark} & \textbf{Multi-Domain Benchmark} & \textbf{Complex Code Benchmark} \\ \midrule

\multicolumn{4}{c}{\textbf{\textit{Train}}} \\ \hline

\textbf{Training Dataset} & 
\makecell{FLAN-v2 training set, \\ 1,000 samples per task} & 
\makecell{A mixed training set includes:\\ Databricks-dolly-15k (general), \\ GenMedGPT, Clinic-10k (medical), \\ Lawyer-Instruct, US-Terms (law), \\ GSM8K training split (math), \\ CodeAlpaca (code)} & 
\makecell{CodeAlpaca} \\ \midrule

\multicolumn{4}{c}{\textbf{\textit{Evaluation}}} \\ \hline

\textbf{Eval Dataset} & 
\makecell{FLAN-v2 eval set, \\ 10 domains, 48 tasks} & 
\makecell{A mixed eval set includes:\\ MMLU, GSM8K, \\ Humaneval} & 
\makecell{DS1000} \\ \midrule


\textbf{Task Types} & 
\makecell{Struct2Text, Translation, \\ 
Commonsense, Sentiment, \\ 
ReadingComp, Closed-QA, \\ 
Coreference, Paraphrasing, NLI} & 
\makecell{Generic language (MMLU), \\ 
Law (MMLU), Medical (MMLU), \\ 
Math (GSM8K), \\ 
Code generation (Humaneval)} & 
\makecell{Pandas, Numpy, \\ 
Matplotlib, Scipy, \\ 
Sklearn, Pytorch, TF} \\ \midrule

\textbf{Evaluation Metrics} & 
\makecell{ROUGE (Struct2Text), \\ 
BLEU (Translation), \\ 
Accuracy (Classification), \\ 
Exact Match (QA)} & 
\makecell{Accuracy (MMLU), \\ 
Step-wise exact match (GSM8K), \\ 
pass@k (k=1, 10) (Humaneval)} & 
\makecell{pass@k (k=1)} \\ \midrule

\textbf{Key Focus} & 
\makecell{Multi-task learning, \\ 
Cross-task knowledge sharing} & 
\makecell{Cross-domain knowledge integration, \\ 
Complex reasoning, \\ 
Code synthesis} & 
\makecell{Domain-specific adaptation, \\ 
Precise code generation} \\ \midrule

\end{tabular}%
}
\end{table}

\subsection{FLAN-v2 Subset}
FLAN-v2 is a large-scale collection of instruction-following tasks designed to evaluate the generalization abilities of language models across various domains. The dataset includes a broad range of tasks, such as question answering, summarization, translation, and reasoning. Due to its heterogeneous nature, FLAN-v2 is especially well-suited for evaluating multi-task learning, as it enables the assessment of how effectively a model can adapt to and transfer knowledge across different tasks.

For our evaluation, we selected a subset of FLAN-v2 comprising \textbf{10 domains} and \textbf{48 tasks}, with a total of \textbf{2,395 test samples}. This subset provides a balanced representation of various task types and domains, allowing us to rigorously assess SMoRA's ability to handle task conflicts and effectively leverage shared knowledge across related tasks. The specific domains and tasks included in our FLAN-v2 subset are as follows:

\paragraph{Struct-to-Text Conversion.} Tasks that involve generating text from structured data. This includes datasets such as CommonGen, DART, E2ENLG, and WebNLG, which test the model's ability to convert structured inputs into coherent natural language descriptions.

\paragraph{Translation.} Machine translation tasks. We utilize datasets like En-Fr from WMT'14, En-De, En-Tr, En-Ru, En-Fi, En-Ro from WMT'16, and En-Es from Paracrawl to assess the model's proficiency in translating text across multiple languages while preserving meaning and nuances.

\paragraph{Commonsense Reasoning.} Tasks requiring commonsense reasoning, such as understanding everyday scenarios. Datasets like COPA, HellaSwag, PiQA, and StoryCloze are used to evaluate the model's ability to apply physical or scientific principles alongside common sense in reasoning tasks.

\paragraph{Sentiment Analysis.} Sentiment analysis tasks, including binary and multi-class classification. We employ datasets such as IMDB, Sentiment140, SST-2, and Yelp to determine the sentiment polarity (positive or negative) of given texts.

\paragraph{Reading Comprehension.} Reading comprehension tasks, where the model answers questions based on provided passages. Datasets like BoolQ, DROP, MultiRC, OBQA, SQuADv1, and SQuADv2 are used to assess the model's capability to derive answers from contextually relevant information.

\paragraph{Closed-Book Question Answering.} Closed-book question answering tasks, where the model answers questions without access to external knowledge. We use datasets such as ARC, NQ, and TriviaQA to challenge the model's ability to recall and apply general knowledge.

\paragraph{Coreference Resolution.} Coreference resolution tasks, which involve identifying expressions that refer to the same entity in a text. Datasets like DPR and WSC273 are used to evaluate the model's understanding of textual context and entity references.

\paragraph{Reading Comprehension with Commonsense.} Reading comprehension tasks with commonsense reasoning, combining text understanding and reasoning. Datasets such as CosmosQA and ReCoRD are used to assess the model's ability to understand and reason beyond the explicit text.

\paragraph{Paraphrase Detection.} Paraphrasing tasks, where the model generates alternative phrasings of given sentences. We use datasets like MRPC, QQP, and Paws Wiki to evaluate the model's ability to detect and generate semantically equivalent sentences.

\paragraph{Natural Language Inference.} Natural Language Inference (NLI) tasks, which involve determining the logical relationship (entailment, contradiction, or neutral) between pairs of sentences. Datasets such as ANLI, CB, MNLI, QNLI, SNLI, WNLI, and RTE are used to assess the model's ability to infer relationships between sentences.

To train each model, we utilized the training set of FLAN-v2. For each task, we randomly selected 1,000 samples and combined them to create a multi-task training dataset.

\subsection{Multi-Domain Benchmark}
We construct a multi-domain evaluation dataset encompassing three major benchmarks: \textbf{MMLU}, \textbf{GSM8K}, and \textbf{Humaneval}, designed to comprehensively assess model capabilities across multiple dimensions. The dataset comprises:  
\begin{itemize}  
    \item \textbf{MMLU}: A multiple-choice dataset spanning 57 academic disciplines, covering STEM, humanities, and social sciences, testing models' breadth of knowledge and reasoning abilities.  
    \item \textbf{GSM8K}: A collection of 8.5K grade-school math word problems requiring multi-step reasoning to solve complex mathematical questions.  
    \item \textbf{HumanEval}: A code generation benchmark containing 164 hand-crafted programming problems that evaluate models' ability to generate functionally correct Python code from natural language descriptions.  
\end{itemize}  

This multi-domain benchmark establishes evaluation challenges across three dimensions: MMLU examines cross-disciplinary knowledge integration, GSM8K verifies mathematical-logical reasoning, and Humaneval tests programmatic semantic understanding. By synthesizing these distinct evaluation perspectives, we systematically analyze SMoRA's comprehensive performance in knowledge-intensive tasks, complex problem-solving, and code-generation scenarios.  

\subsection{DS-1000}
DS-1000 is a benchmark specifically designed to evaluate code generation in the data science domain. It consists of \textbf{1,000 coding problems} across \textbf{7 different Python tasks}: Pandas, Numpy, Matplotlib, Scipy, Sklearn, Pytorch, and Tensorflow. Each problem includes a natural language description, and the model is tasked with generating the corresponding code snippet.

\subsection{Evaluation Metrics}  
We employ a fine-grained evaluation metric system tailored to task characteristics:  
\begin{itemize}  
    \item \textbf{FLAN-v2 Subset}: Preserves original evaluation protocols including \textbf{ROUGE} (text generation), \textbf{BLEU} (machine translation), \textbf{accuracy} (classification tasks), and \textbf{exact match} (QA tasks).  
    \item \textbf{Multi-Domain Benchmark}:  
    \begin{itemize}  
        \item \textbf{MMLU}: Strict accuracy measuring exact match between predictions and ground-truth answers.  
        \item \textbf{GSM8K}: Step-wise exact match requiring both correct reasoning chains and final answers.  
        \item \textbf{Humaneval}: Pass@k metric (recommended k=1,10) validating functional correctness through unit tests.  
    \end{itemize}  
    \item \textbf{DS-1000}: We use \textbf{pass@1} to evaluate code generation performance.
\end{itemize}  

These metrics allow us to quantify the effectiveness of SMoRA in both multi-task learning and domain-specific adaptation scenarios, providing a holistic view of its capabilities.

\begin{table}[h!]
\centering
\caption{\textbf{Accuracy Across Different Python Libraries in DS-1000.} The best performance for each task is highlighted in bold, and the second-best performance is underlined. The performance of our proposed SMoRA is highlighted in gray.}
\label{tab:results_ds1000}
\resizebox{.85\linewidth}{!}{%
\begin{tabular}{lccccccc|c}
\toprule
\textbf{Method}       & \textbf{Pandas}  & \textbf{Numpy}  & \textbf{Tensorflow} & \textbf{Scipy} & \textbf{Sklearn} & \textbf{Pytorch} & \textbf{Matplotlib} & \textbf{Average} \\
\midrule
    \textbf{LoRA-64}     & \underline{6.19}  & 5.45  & 13.3  & \underline{7.55}  & 6.96  & 0.00  & 45.16 & 12.09  \\
    \textbf{MoE-Soft}    & 5.84   & 5.91  & \underline{15.56}  & \textbf{9.43}   & 6.96   & \underline{1.47}   & 46.45  & 13.09  \\
    \textbf{MoE-Top1}    & \underline{6.19}  & 5.00  & \textbf{17.78} & \underline{7.55}  & \underline{8.70}  & \textbf{2.94}  & 47.74 & 13.71  \\
    \textbf{MoE-Top2}    & \textbf{6.53} & 5.45  & \textbf{17.78} & \underline{7.55}  & 7.83  & 0.00  & \textbf{51.61} & 13.82  \\
    \textbf{HydraLoRA}   & 5.50  & 5.00  & \textbf{17.78}  & \underline{7.55} & 6.96 & \underline{1.47} & 48.39 & 13.24  \\
    \textbf{MoSLoRA}     & \underline{6.19}  & \textbf{6.82} & 11.11  & \underline{7.55} & 7.83 & 0.00  & 43.23 & 11.82  \\
    \textbf{SMEAR}       & \underline{6.19}  & 5.00  & 13.3  & \underline{7.55}  & 7.83  & 0.00  & 45.16 & 11.82  \\ \midrule
    \textbf{SMoRA}       & \cellcolor[gray]{0.9}\underline{6.19} & \cellcolor[gray]{0.9}\underline{6.36} & \cellcolor[gray]{0.9}\underline{15.56} & \cellcolor[gray]{0.9}6.60 & \cellcolor[gray]{0.9}\textbf{12.17} & \cellcolor[gray]{0.9}\underline{1.47} & \cellcolor[gray]{0.9}\underline{49.68} & \cellcolor[gray]{0.9}\textbf{14.00}  \\
\bottomrule
\end{tabular}}
    \vspace{-4mm}
\end{table}

\section{Main Results on DS-1000}
\label{app:python_results}
Table~\ref{tab:results_ds1000} shows the results on the DS-1000 dataset to demonstrate the effectiveness when adapting to specific downstream tasks such as code generation. The results also show that: (1) LoRA with full rank activation performs poorly on the code dataset due to the impact of data heterogeneity; (2) Methods like MoE, which decouple parameters, can alleviate task interference to some extent, and their performance improves as the number of activated experts increases; (3) SMoRA, with only 8 rank parameters activated, achieves the best performance, as its fine-grained parameter decoupling effectively avoids task interference while retaining shared knowledge, thus achieving optimal results on downstream tasks.

\section{Implementation Details of Baseline Methods}
\label{app:baseline_methods}
In this section, we provide a detailed explanation of the baseline methods used for comparison with the proposed SMoRA. Each method is described with its core idea, mathematical formulation, and specific implementation details. For all methods, we use \textbf{Kaiming initialization} for matrix $\mA$ and \textbf{zero initialization} for matrix $\mB$ in the LoRA modules, unless otherwise specified. This ensures that the initial update $\Delta W = B \cdot A = 0$, preserving the pre-trained model's performance at the start of training.

\subsection{LoRA with Different Ranks}
\textbf{Idea}:  
LoRA reduces the number of trainable parameters by injecting low-rank decomposition matrices $\mA$ and $\mB$ into the pre-trained model's weights. The key idea is that the weight updates during fine-tuning lie in a low-dimensional subspace, allowing efficient adaptation with minimal parameters.

\textbf{Formulation}:  
For a pre-trained weight matrix $W_0$, the update is constrained to a low-rank decomposition:  
\[
\Delta W = B \cdot A,
\]  
where $B \in \mathbb{R}^{d \times r}$ and $A \in \mathbb{R}^{r \times d}$, with $r \ll d$. The updated weight matrix becomes:  
\[
W = W_0 + \Delta W = W_0 + B \cdot A,
\]  
Where $r$ is the rank of the decomposition, controlling the number of trainable parameters.

\textbf{Implementation}:  
We test LoRA with ranks of \textbf{64, 32, 16, and 8}. Matrix $\mA$ is initialized using \textbf{Kaiming initialization}, and matrix $\mB$ is initialized to \textbf{zero}.

\subsection{LoRA MoE with Different Routing Strategies}
LoRA MoE methods introduce multiple LoRA modules (experts) and use a router to dynamically select and combine the outputs of these experts. This approach mitigates task interference by isolating task-specific adaptations while maintaining parameter efficiency.

\textbf{Formulation}:  
Let the parameter of $i$-th LoRA expert denotes as $\{A_i, B_i\}$ and $G$ denote the router. The output of the MoE system can be expressed as:  
\[
y = W_0 x + \sum_{i=1}^N G(\vx)_i B_iA_ix,
\]  
where $G(\vx)_i$ is the gating score for the $i$-th expert, and $N$ is the number of active experts.

We implement two variants of the MoE routing mechanism:  
\begin{itemize}
    \item \textbf{Dense Gating}: The router consists of a dense layer with trainable parameters $W_g$. The gating scores are computed using a softmax function:  
    \[
    g_i = G(\vx)_i = \text{softmax}(W_g^T x).
    \]  
    Dense gating is used for the \textbf{Soft Routing} strategy, where the outputs of multiple experts are weighted and combined based on their gating scores.
    \item \textbf{Top-1 Gating}: To maintain sparsity during training, we leverage the Gumbel softmax trick. The router can be written as:  
    \[
    \hat{G}(\vx)_i = \frac{\exp((\log(G(\vx)_i) + g_i) / \tau)}{\sum_{j=1}^k \exp((\log(G(\vx)_j) + g_j) / \tau)},
    \]  
    where $g_i \sim \text{Gumbel}(0,1)$ and $\tau$ is the temperature. Sparse gating is used for \textbf{Top-1} routing, where only the top expert is activated for each input.
    \item \textbf{Top-k Gating}: For the Top-k routing strategy, we use a routing mechanism similar to Mixtral~\cite{jiang2024mixtral}, which selects the top-k experts based on their gating scores and combines their outputs accordingly. Specifically, the gating function is defined as: $g(\vx) = \mathrm{Softmax}(\mathrm{TopK}(x W_g))$, where only the top-k experts are considered for each input.
\end{itemize}

\textbf{Implementation}:  
We implement LoRA MoE with \textbf{8 experts}, each having a rank of \textbf{8}. The router employs one of three strategies—top-1, top-2, or soft routing—depending on the input, ensuring efficient adaptation to specific tasks.

\subsection{SMEAR}
\textbf{Idea}:  
SMEAR aggregates LoRA modules at the parameter level rather than at the output level. This allows for more flexible adaptation by combining the parameters of multiple LoRA modules based on gating scores.

\textbf{Formulation}:  
The aggregated parameters $\Theta_{SMEAR}$ are computed as:  
\[
\Theta_{SMEAR} = \sum_{i=1}^k G(\vx)_i \Theta_i,
\]  
where $\Theta_i$ denotes the parameters of the $i$-th LoRA module, and $G(\vx)_i$ is the gating score for the $i$-th module.

\textbf{Implementation}:  
We adopt the same settings as the MoE methods, with \textbf{8 experts}, each with a rank of \textbf{8}. The gating scores are learned during training, enabling an adaptive combination of LoRA parameters.

\subsection{HydraLoRA}
\textbf{Idea}:  
HydraLoRA introduced an asymmetric LoRA structure. The core idea is to decompose the traditional LoRA into a shared central matrix $\mA$ and multiple distinct matrices $\mB_i$, enabling both knowledge sharing and functional specialization.

\textbf{Formulation}:  
HydraLoRA introduces multiple independent matrices $\mB_i$ that share a central matrix $\mA$. Each $\mB_i$ is modulated by a contribution weight $\omega_i$, allowing for task-specific adaptations while maintaining shared knowledge through $\mA$. The forward process is modified as:  
\[
y = W_0 x +  \sum_{i=1}^N G(\vx)_i B_iA x,
\]  
where $B_i \in \mathbb{R}^{d \times r}$ are independent low-rank matrices for task-specific adaptations, $A \in \mathbb{R}^{r \times k}$ is the shared central matrix for knowledge sharing, $\omega_i$ are contribution weights that modulate the influence of each $\mB_i$, and $N$ is the number of distinct $\mB$ matrices.

\textbf{Implementation}:  
We configure HydraLoRA with \textbf{4 distinct $\mB$ matrices}, each with a rank of \textbf{8}. The shared central matrix $\mA$ is initialized using \textbf{Kaiming initialization}, and the distinct matrices $\mB_i$ are initialized to \textbf{zero}. 

\subsection{MoSLoRA}
\textbf{Idea}:  
MoSLoRA introduces a trainable mixer \(\mathbf{W}\) to fuse subspace information more effectively, enhancing the model's flexibility and expressive power. Unlike traditional LoRA, which uses a fixed identity matrix for subspace fusion, MoSLoRA employs a trainable mixer to enable richer interactions between subspaces.

\textbf{Formulation}:  
The core idea of MoSLoRA is to replace the fixed mixer in traditional LoRA with a trainable matrix \({W} \in \mathbb{R}^{r \times r}\):
\[
W = W_0 + \Delta W = W_0 + B  W  A,
\]
where \({W}\) is a trainable matrix that learns to combine the subspaces in a data-driven manner, allowing for richer interactions between the low-rank adaptations.

\textbf{Implementation}:  
We configure MoSLoRA with a rank of \textbf{64}. The trainable mixer \({W} \in \mathbb{R}^{r \times r}\) is initialized using \textbf{Kaiming initialization}. 

\section{TVM Implementation of \textit{indexed\_matmul}}
\label{app:tvm}
TVM~\cite{chen2018tvm} is an end-to-end deep learning compiler stack designed to optimize machine learning workloads across diverse hardware backends. It enables efficient computation by providing automated tensor optimization techniques such as operation fusion, memory layout transformations, and parallelization, making it particularly suitable for high-performance deep learning tasks.

In our work, we implemented a custom CUDA kernel, \textit{indexed\_matmul}, using TVM to accelerate sparse matrix computations in SMoRA. The key innovation of \textit{indexed\_matmul} lies in its dynamic handling of sparse operations. Instead of performing a full matrix multiplication, the kernel leverages the top-K indices from the gating function to dynamically extract and compute only the relevant rows and columns of the LoRA matrices. This approach reduces computational and memory overhead by avoiding redundant calculations.

To further optimize the kernel, we utilized several TVM-specific GPU optimizations:
\begin{itemize}
    \item \textbf{Thread and Block-Level Parallelism}: Using TVM's scheduling primitives, we split and reordered computation across blocks and threads, enabling efficient parallel execution of sparse operations.
    \item \textbf{Reduction Fusion}: The reduction axis in the sparse matrix computation was fused and optimized to minimize intermediate memory usage and improve computational efficiency.
    \item \textbf{Dynamic Index-Based Computation}: By focusing on the non-zero indices identified by the gating mechanism, the kernel avoids loading or processing irrelevant data, further improving execution speed and memory efficiency.
\end{itemize}

\section{Full Results on FLAN-v2}
\label{sec:full_res}
In this section, we present the full results of our experiments conducted on FLAN-v2, covering 10 domains and 48 tasks. The results for the two base models, Llama-\{7b,13b\}, are shown in Tab.\ref{tab:flan-full-7b} and Tab.\ref{tab:flan-full-13b}.

\begin{table*}[h!]
\centering
\caption{\textbf{Full Performance of Llama2-7b on FLAN-v2.}}
\label{tab:flan-full-7b}
\resizebox{\linewidth}{!}{%
\begin{tabular}{lccccccccccc}
\toprule
\textbf{Task}  & \textbf{SMoRA-64-8} & \textbf{LoRA-64} & \textbf{LoRA-32} & \textbf{LoRA-16} & \textbf{LoRA-8} & \textbf{MoE-Top1} & \textbf{MoE-Top2} & \textbf{MoE-Soft} & \textbf{SMEAR} & \textbf{HydraLoRA} & \textbf{MosLoRA}\\
\midrule

 \textbf{Translation} & & & & & & & & & & & \\
 WMT'16-tren & 2.0 & 2.2 & 2.8 & 3.7 & 3.9 & 3.4 & 2.6 & 3.4 & 2.9 & 3.3 & 2.6 \\
 WMT'16-roen & 14.1 & 14.0 & 13.8 & 14.0 & 15.5 & 13.7 & 13.9 & 13.3 & 13.3 & 14.1 & 14.4 \\
 WMT'16-fien & 7.4 & 8.2 & 7.2 & 8.4 & 7.7 & 8.5 & 8.0 & 8.1 & 8.4 & 9.3 & 8.2 \\
 WMT'16-ruen & 13.1 & 11.4 & 11.8 & 12.1 & 10.6 & 11.2 & 12.2 & 11.9 & 13.1 & 11.8 & 13.2 \\
 WMT'16-deen & 17.5 & 16.9 & 19.3 & 18.3 & 18.6 & 18.2 & 17.0 & 20.3 & 19.9 & 19.2 & 19.6 \\
 Paracrawl-enes & 24.3 & 25.9 & 23.1 & 22.4 & 23.3 & 21.7 & 22.2 & 23.0 & 22.8 & 22.0 & 25.5 \\
 WMT'16-csen & 8.6 & 8.1 & 7.8 & 7.8 & 8.6 & 6.8 & 8.5 & 7.8 & 7.8 & 8.7 & 7.5 \\
 WMT'14-enfr & 17.3 & 18.3 & 18.0 & 17.8 & 18.4 & 17.5 & 18.2 & 17.2 & 17.9 & 18.0 & 17.5 \\
\midrule

 \textbf{Read.Comp.W:Commonsense } & & & & & & & & & & \\
 CosmosQa & 72.0 & 68.0 & 68.0 & 64.0 & 64.0 & 60.0 & 66.0 & 68.0 & 68.0 & 64.0 & 68.0 \\
 record & 70.0 & 66.0 & 66.0 & 58.0 & 42.0 & 48.0 & 60.0 & 52.0 & 54.0 & 44.0 & 64.0 \\
\midrule

 \textbf{Reading Comp} & & & & & & & & & & & \\
 OBQA & 68.0 & 68.0 & 70.0 & 66.0 & 62.0 & 62.0 & 68.0 & 68.0 & 70.0 & 62.0 & 68.0 \\
 drop & 14.0 & 12.0 & 10.0 & 20.0 & 16.0 & 18.0 & 16.0 & 26.0 & 20.0 & 18.0 & 14.0 \\
 SQuADv1 & 66.0 & 66.0 & 62.0 & 58.0 & 52.0 & 58.0 & 62.0 & 60.0 & 62.0 & 58.0 & 60.0 \\
 BoolQ & 84.0 & 74.0 & 78.0 & 78.0 & 76.0 & 78.0 & 80.0 & 76.0 & 78.0 & 74.0 & 74.0 \\
 SQuADv2 & 18.0 & 16.0 & 16.0 & 14.0 & 12.0 & 16.0 & 14.0 & 12.0 & 14.0 & 14.0 & 16.0 \\
 MultiRC & 66.0 & 56.0 & 56.0 & 58.0 & 48.0 & 56.0 & 62.0 & 52.0 & 56.0 & 54.0 & 56.0 \\
\midrule

 \textbf{NLI} & & & & & & & & & & & \\
 ANLI-r3 & 40.0 & 36.0 & 30.0 & 36.0 & 30.0 & 34.0 & 36.0 & 38.0 & 40.0 & 30.0 & 44.0 \\
 SNLI & 84.0 & 78.0 & 86.0 & 82.0 & 90.0 & 62.0 & 80.0 & 82.0 & 80.0 & 92.0 & 88.0 \\
 ANLI-r2 & 52.0 & 36.0 & 38.0 & 42.0 & 42.0 & 38.0 & 38.0 & 38.0 & 38.0 & 44.0 & 36.0 \\
 CB & 97.8 & 100.0 & 95.6 & 95.6 & 93.3 & 95.6 & 93.3 & 91.1 & 91.1 & 91.1 & 100.0 \\
 MNLI-mm & 82.0 & 72.0 & 68.0 & 80.0 & 80.0 & 42.0 & 68.0 & 72.0 & 78.0 & 84.0 & 78.0 \\
 MNLI-m & 82.0 & 80.0 & 82.0 & 86.0 & 80.0 & 58.0 & 76.0 & 82.0 & 80.0 & 82.0 & 90.0 \\
 WNLI & 62.0 & 64.0 & 64.0 & 54.0 & 52.0 & 50.0 & 66.0 & 64.0 & 64.0 & 58.0 & 70.0 \\
 ANLI-r1 & 48.0 & 58.0 & 50.0 & 48.0 & 36.0 & 40.0 & 50.0 & 50.0 & 42.0 & 40.0 & 56.0 \\
 QNLI & 88.0 & 88.0 & 84.0 & 78.0 & 70.0 & 72.0 & 88.0 & 74.0 & 80.0 & 82.0 & 88.0 \\
 RTE & 82.0 & 80.0 & 84.0 & 86.0 & 84.0 & 82.0 & 86.0 & 84.0 & 84.0 & 80.0 & 80.0 \\
\midrule

 \textbf{Sentiment} & & & & & & & & & & & \\
 Yelp & 98.0 & 98.0 & 98.0 & 98.0 & 98.0 & 98.0 & 98.0 & 98.0 & 98.0 & 98.0 & 98.0 \\
 SST-2 & 100.0 & 100.0 & 100.0 & 100.0 & 100.0 & 98.0 & 100.0 & 100.0 & 98.0 & 100.0 & 100.0 \\
 sentiment140 & 72.0 & 70.0 & 70.0 & 74.0 & 70.0 & 70.0 & 72.0 & 70.0 & 72.0 & 70.0 & 70.0 \\
 IMDB & 94.0 & 92.0 & 92.0 & 94.0 & 94.0 & 94.0 & 92.0 & 94.0 & 94.0 & 94.0 & 94.0 \\
\midrule

 \textbf{Paraphrase} & & & & & & & & & & & \\
 STSB & 40.0 & 42.0 & 36.0 & 32.0 & 22.0 & 36.0 & 30.0 & 40.0 & 36.0 & 34.0 & 38.0 \\
 Paws Wiki & 80.0 & 80.0 & 76.0 & 70.0 & 62.0 & 64.0 & 78.0 & 64.0 & 68.0 & 68.0 & 78.0 \\
 QQP & 80.0 & 84.0 & 80.0 & 74.0 & 72.0 & 72.0 & 80.0 & 76.0 & 74.0 & 68.0 & 80.0 \\
 MRPC & 74.0 & 74.0 & 72.0 & 72.0 & 68.0 & 70.0 & 72.0 & 70.0 & 72.0 & 72.0 & 74.0 \\
\midrule

 \textbf{Struct to Text} & & & & & & & & & & & \\
 WebNLG 	\textsuperscript{Rouge-1} & 68.8 & 66.6 & 65.5 & 62.6 & 60.9 & 64.3 & 63.9 & 64.2 & 61.6 & 62.2 & 67.6 \\
 WebNLG 	\textsuperscript{Rouge-2} & 45.5 & 42.0 & 41.0 & 36.6 & 35.3 & 38.5 & 38.7 & 38.9 & 35.4 & 37.5 & 43.4 \\
 WebNLG 	\textsuperscript{Rouge-l} & 61.4 & 60.2 & 58.2 & 56.2 & 54.8 & 57.9 & 57.9 & 57.9 & 54.7 & 56.0 & 60.6 \\
 E2ENLG 	\textsuperscript{Rouge-1} & 65.8 & 65.2 & 65.2 & 65.4 & 65.8 & 65.4 & 65.1 & 65.6 & 64.5 & 64.8 & 66.0 \\
 E2ENLG 	\textsuperscript{Rouge-2} & 38.6 & 37.6 & 38.0 & 38.3 & 38.3 & 38.4 & 38.0 & 38.9 & 38.2 & 37.9 & 38.2 \\
 E2ENLG 	\textsuperscript{Rouge-l} & 55.3 & 54.6 & 55.5 & 55.9 & 56.4 & 57.1 & 56.0 & 56.7 & 55.4 & 55.1 & 55.6 \\
 CommonGen 	\textsuperscript{Rouge-1} & 45.3 & 46.2 & 46.4 & 45.2 & 44.3 & 45.1 & 45.8 & 45.9 & 44.7 & 44.4 & 45.6 \\
 CommonGen 	\textsuperscript{Rouge-2} & 17.7 & 19.6 & 19.1 & 18.0 & 17.5 & 17.1 & 17.4 & 18.3 & 17.0 & 16.3 & 17.9 \\
 CommonGen 	\textsuperscript{Rouge-l} & 41.3 & 41.5 & 42.4 & 41.0 & 40.8 & 42.1 & 41.8 & 42.4 & 40.6 & 39.1 & 41.5 \\
 DART 	\textsuperscript{Rouge-1} & 70.9 & 71.6 & 70.9 & 66.2 & 65.7 & 66.1 & 69.8 & 67.6 & 67.5 & 68.0 & 69.6 \\
 DART 	\textsuperscript{Rouge-2} & 48.5 & 49.8 & 48.5 & 43.0 & 39.8 & 41.8 & 47.3 & 45.1 & 43.6 & 44.2 & 47.0 \\
 DART 	\textsuperscript{Rouge-l} & 63.6 & 63.5 & 63.8 & 59.7 & 58.4 & 59.7 & 62.4 & 61.2 & 61.3 & 61.3 & 61.7 \\
\midrule

 \textbf{Coreference} & & & & & & & & & & & \\
 DPR & 72.0 & 80.0 & 66.0 & 66.0 & 68.0 & 64.0 & 62.0 & 64.0 & 70.0 & 72.0 & 80.0 \\
 WSC & 46.0 & 42.0 & 44.0 & 48.0 & 44.0 & 50.0 & 42.0 & 50.0 & 48.0 & 52.0 & 44.0 \\
\midrule

 \textbf{Closed Book QA} & & & & & & & & & & & \\
 NQ & 22.0 & 16.0 & 20.0 & 24.0 & 22.0 & 20.0 & 22.0 & 22.0 & 22.0 & 20.0 & 18.0 \\
 TriviaQa & 60.0 & 62.0 & 58.0 & 56.0 & 54.0 & 58.0 & 60.0 & 58.0 & 54.0 & 58.0 & 58.0 \\
 ARC-e & 86.0 & 84.0 & 84.0 & 82.0 & 76.0 & 78.0 & 88.0 & 82.0 & 80.0 & 80.0 & 86.0 \\
 ARC-c & 54.0 & 56.0 & 54.0 & 48.0 & 46.0 & 48.0 & 56.0 & 50.0 & 48.0 & 48.0 & 56.0 \\
\midrule

 \textbf{Commonsense} & & & & & & & & & & & \\
 PIQA & 38.0 & 40.0 & 42.0 & 40.0 & 42.0 & 36.0 & 38.0 & 38.0 & 40.0 & 40.0 & 42.0 \\
 HellaSwag & 52.0 & 52.0 & 54.0 & 44.0 & 40.0 & 40.0 & 50.0 & 40.0 & 40.0 & 42.0 & 52.0 \\
 COPA & 82.0 & 78.0 & 78.0 & 76.0 & 76.0 & 76.0 & 76.0 & 78.0 & 74.0 & 78.0 & 74.0 \\
 StoryCloze & 96.0 & 94.0 & 90.0 & 94.0 & 88.0 & 90.0 & 92.0 & 92.0 & 90.0 & 92.0 & 92.0 \\

\bottomrule
\end{tabular}%
}
\end{table*}

\begin{table*}[h!]
\centering
\caption{\textbf{Full Performance of Llama2-13b on FLAN-v2.}}
\label{tab:flan-full-13b}
\resizebox{\linewidth}{!}{%
\begin{tabular}{lccccccccccc}
\toprule 
\textbf{Task}  & \textbf{SMoRA-64-8} & \textbf{LoRA-64} & \textbf{LoRA-32} & \textbf{LoRA-16} & \textbf{LoRA-8} & \textbf{MoE-Top1} & \textbf{MoE-Top2} & \textbf{MoE-Soft} & \textbf{SMEAR} & \textbf{HydraLoRA} & \textbf{MosLoRA}\\
\midrule

 \textbf{Translation} & & & & & & & & & & & \\
 WMT'16-tren & 3.5 & 3.7 & 4.2 & 2.9 & 3.0 & 3.2 & 2.8 & 3.3 & 2.9 & 3.2 & 4.0 \\
 WMT'16-roen & 15.4 & 15.2 & 14.2 & 14.4 & 15.7 & 15.6 & 16.1 & 15.4 & 15.3 & 14.7 & 14.8 \\
 WMT'16-fien & 7.9 & 7.5 & 7.6 & 7.1 & 8.6 & 7.7 & 8.3 & 7.9 & 8.4 & 7.5 & 8.0 \\
 WMT'16-ruen & 10.6 & 13.0 & 12.3 & 13.5 & 11.2 & 9.8 & 10.9 & 11.0 & 12.6 & 12.8 & 13.0 \\
 WMT'16-deen & 19.6 & 19.9 & 19.0 & 17.4 & 17.2 & 18.1 & 19.3 & 17.7 & 17.9 & 18.2 & 17.8 \\
 Paracrawl-enes & 26.4 & 27.5 & 27.6 & 26.6 & 27.7 & 26.0 & 26.5 & 26.4 & 28.6 & 28.3 & 27.0 \\
 WMT'16-csen & 10.9 & 10.6 & 11.4 & 13.4 & 12.8 & 10.3 & 11.7 & 10.8 & 12.0 & 11.5 & 11.3 \\
 WMT'14-enfr & 19.9 & 19.7 & 19.7 & 21.4 & 20.8 & 19.6 & 20.0 & 20.7 & 19.6 & 20.2 & 19.0 \\
\midrule

 \textbf{Read.Comp.W:Commonsense } & & & & & & & & & & \\
 CosmosQa & 82.0 & 82.0 & 82.0 & 80.0 & 82.0 & 80.0 & 80.0 & 78.0 & 80.0 & 78.0 & 86.0 \\
 record & 64.0 & 60.0 & 54.0 & 50.0 & 50.0 & 46.0 & 58.0 & 54.0 & 54.0 & 58.0 & 62.0 \\
\midrule

 \textbf{Reading Comp} & & & & & & & & & & & \\
 OBQA & 72.0 & 72.0 & 70.0 & 68.0 & 66.0 & 64.0 & 68.0 & 64.0 & 66.0 & 66.0 & 72.0 \\
 drop & 26.0 & 22.0 & 20.0 & 16.0 & 16.0 & 18.0 & 24.0 & 20.0 & 18.0 & 20.0 & 22.0 \\
 SQuADv1 & 62.0 & 60.0 & 60.0 & 56.0 & 54.0 & 58.0 & 62.0 & 60.0 & 62.0 & 54.0 & 60.0 \\
 BoolQ & 88.0 & 90.0 & 88.0 & 88.0 & 90.0 & 84.0 & 86.0 & 86.0 & 88.0 & 88.0 & 88.0 \\
 SQuADv2 & 20.0 & 16.0 & 18.0 & 16.0 & 18.0 & 14.0 & 12.0 & 16.0 & 14.0 & 18.0 & 18.0 \\
 MultiRC & 64.0 & 62.0 & 60.0 & 60.0 & 64.0 & 66.0 & 58.0 & 66.0 & 64.0 & 64.0 & 66.0 \\
\midrule

 \textbf{NLI} & & & & & & & & & & & \\
 ANLI-r3 & 50.0 & 50.0 & 50.0 & 48.0 & 50.0 & 44.0 & 56.0 & 48.0 & 50.0 & 44.0 & 48.0 \\
 SNLI & 90.0 & 90.0 & 86.0 & 84.0 & 76.0 & 84.0 & 90.0 & 86.0 & 86.0 & 82.0 & 90.0 \\
 ANLI-r2 & 44.0 & 48.0 & 50.0 & 46.0 & 40.0 & 50.0 & 50.0 & 52.0 & 54.0 & 56.0 & 52.0 \\
 CB & 86.7 & 93.3 & 95.6 & 93.3 & 91.1 & 93.3 & 97.8 & 95.6 & 93.3 & 91.1 & 95.6 \\
 MNLI-mm & 88.0 & 96.0 & 94.0 & 96.0 & 94.0 & 96.0 & 94.0 & 98.0 & 98.0 & 98.0 & 92.0 \\
 MNLI-m & 86.0 & 90.0 & 90.0 & 88.0 & 82.0 & 84.0 & 86.0 & 84.0 & 92.0 & 80.0 & 90.0 \\
 WNLI & 74.0 & 64.0 & 68.0 & 66.0 & 64.0 & 66.0 & 66.0 & 68.0 & 66.0 & 66.0 & 58.0 \\
 ANLI-r1 & 58.0 & 50.0 & 46.0 & 48.0 & 54.0 & 48.0 & 50.0 & 46.0 & 54.0 & 54.0 & 52.0 \\
 QNLI & 82.0 & 82.0 & 84.0 & 80.0 & 76.0 & 86.0 & 80.0 & 84.0 & 84.0 & 80.0 & 80.0 \\
 RTE & 92.0 & 84.0 & 82.0 & 78.0 & 80.0 & 82.0 & 80.0 & 80.0 & 80.0 & 80.0 & 82.0 \\
\midrule

 \textbf{Sentiment} & & & & & & & & & & & \\
 Yelp & 98.0 & 98.0 & 98.0 & 98.0 & 98.0 & 98.0 & 98.0 & 98.0 & 98.0 & 98.0 & 98.0 \\
 SST-2 & 100.0 & 100.0 & 98.0 & 98.0 & 100.0 & 100.0 & 100.0 & 100.0 & 100.0 & 100.0 & 100.0 \\
 sentiment140 & 72.0 & 70.0 & 72.0 & 76.0 & 76.0 & 74.0 & 74.0 & 76.0 & 72.0 & 78.0 & 70.0 \\
 IMDB & 94.0 & 94.0 & 94.0 & 94.0 & 96.0 & 94.0 & 94.0 & 94.0 & 92.0 & 94.0 & 94.0 \\
\midrule

 \textbf{Paraphrase} & & & & & & & & & & & \\
 STSB & 30.0 & 36.0 & 32.0 & 36.0 & 36.0 & 38.0 & 32.0 & 28.0 & 34.0 & 34.0 & 34.0 \\
 Paws Wiki & 84.0 & 82.0 & 76.0 & 78.0 & 76.0 & 78.0 & 78.0 & 82.0 & 80.0 & 82.0 & 84.0 \\
 QQP & 82.0 & 80.0 & 88.0 & 82.0 & 82.0 & 76.0 & 86.0 & 84.0 & 84.0 & 82.0 & 82.0 \\
 MRPC & 74.0 & 70.0 & 66.0 & 66.0 & 64.0 & 74.0 & 70.0 & 68.0 & 70.0 & 70.0 & 66.0 \\
\midrule

 \textbf{Struct to Text} & & & & & & & & & & & \\
 WebNLG \textsuperscript{Rouge-1} & 69.2 & 68.5 & 66.8 & 65.7 & 63.9 & 64.1 & 67.1 & 64.7 & 66.4 & 64.6 & 68.2 \\
 WebNLG \textsuperscript{Rouge-2} & 46.9 & 45.6 & 44.3 & 42.8 & 40.7 & 40.5 & 43.9 & 40.8 & 42.1 & 41.7 & 46.4 \\
 WebNLG \textsuperscript{Rouge-l} & 62.1 & 60.8 & 60.0 & 59.9 & 58.3 & 58.5 & 61.6 & 60.0 & 59.3 & 59.1 & 62.3 \\
 E2ENLG \textsuperscript{Rouge-1} & 67.1 & 66.5 & 66.2 & 65.4 & 65.3 & 65.8 & 65.9 & 66.3 & 66.8 & 65.4 & 66.4 \\
 E2ENLG \textsuperscript{Rouge-2} & 39.1 & 38.7 & 37.7 & 38.0 & 37.3 & 39.0 & 38.1 & 38.6 & 38.6 & 37.3 & 38.4 \\
 E2ENLG \textsuperscript{Rouge-l} & 55.3 & 55.9 & 56.2 & 55.8 & 55.0 & 56.6 & 55.3 & 55.2 & 57.0 & 55.5 & 54.0 \\
 CommonGen \textsuperscript{Rouge-1} & 46.5 & 45.7 & 46.6 & 48.0 & 45.8 & 47.0 & 47.8 & 47.1 & 46.1 & 46.2 & 49.1 \\
 CommonGen \textsuperscript{Rouge-2} & 18.8 & 20.0 & 21.5 & 20.1 & 15.8 & 17.4 & 20.9 & 18.8 & 16.1 & 17.0 & 22.0 \\
 CommonGen \textsuperscript{Rouge-l} & 42.4 & 41.3 & 42.3 & 44.0 & 41.5 & 43.9 & 43.2 & 43.4 & 41.6 & 42.0 & 45.1 \\
 DART \textsuperscript{Rouge-1} & 71.3 & 72.9 & 70.7 & 67.7 & 65.6 & 67.9 & 70.2 & 70.8 & 69.8 & 68.4 & 71.6 \\
 DART \textsuperscript{Rouge-2} & 49.4 & 50.6 & 48.2 & 43.8 & 42.7 & 44.5 & 46.4 & 47.5 & 46.9 & 44.6 & 49.8 \\
 DART \textsuperscript{Rouge-l} & 63.1 & 64.0 & 63.1 & 61.4 & 59.6 & 62.8 & 62.3 & 64.3 & 63.9 & 62.8 & 64.5 \\
\midrule

 \textbf{Coreference} & & & & & & & & & & & \\
 DPR & 86.0 & 80.0 & 80.0 & 72.0 & 72.0 & 74.0 & 78.0 & 68.0 & 70.0 & 70.0 & 76.0 \\
 WSC & 50.0 & 54.0 & 52.0 & 54.0 & 52.0 & 58.0 & 50.0 & 56.0 & 50.0 & 56.0 & 52.0 \\
\midrule

 \textbf{Closed Book QA} & & & & & & & & & & & \\
 NQ & 28.0 & 22.0 & 22.0 & 22.0 & 20.0 & 20.0 & 28.0 & 26.0 & 22.0 & 18.0 & 24.0 \\
 TriviaQa & 64.0 & 64.0 & 62.0 & 68.0 & 68.0 & 66.0 & 66.0 & 68.0 & 68.0 & 68.0 & 62.0 \\
 ARC-e & 94.0 & 94.0 & 94.0 & 94.0 & 92.0 & 94.0 & 94.0 & 94.0 & 94.0 & 92.0 & 94.0 \\
 ARC-c & 60.0 & 58.0 & 56.0 & 52.0 & 48.0 & 52.0 & 54.0 & 52.0 & 52.0 & 54.0 & 58.0 \\
\midrule

 \textbf{Commonsense} & & & & & & & & & & & \\
 PIQA & 46.0 & 44.0 & 42.0 & 40.0 & 44.0 & 42.0 & 48.0 & 42.0 & 42.0 & 42.0 & 44.0 \\
 HellaSwag & 50.0 & 44.0 & 46.0 & 50.0 & 46.0 & 48.0 & 38.0 & 46.0 & 52.0 & 50.0 & 40.0 \\
 COPA & 86.0 & 88.0 & 86.0 & 86.0 & 86.0 & 84.0 & 86.0 & 84.0 & 84.0 & 84.0 & 88.0 \\
 StoryCloze & 98.0 & 98.0 & 98.0 & 96.0 & 100.0 & 100.0 & 98.0 & 96.0 & 98.0 & 100.0 & 98.0 \\

\bottomrule
\end{tabular}%
}
\end{table*}

\section{Difference between SMoRA and MoSLoRA}
Although the forward modeling of SMoRA is given by  $y = W_0 x + B G(\vx) A x$ and the forward of MoSLoRA can be written as $y = W_0 x + B W A x$, they may appear similar in form, but they are fundamentally different. Firstly, in MoSLoRA, the mixture matrix is fixed after training, unlike SMoRA, where expert selection is based on the current token. This leads to MoSLoRA introducing more parameters and losing the flexibility of dynamic routing. Additionally, MoSLoRA activates and mixes all the LoRA ranks, whereas we adopt a dynamic sparse activation approach to efficiently decouple parameters. This allows us to achieve better performance with fewer active parameters.

\section{Visualization of Routing Distribution Under Load Balancing Strategy Ablation}
\label{app:routing_ablation}
\begin{figure*}[h!]
    \centering
\includegraphics[width=.95\linewidth]{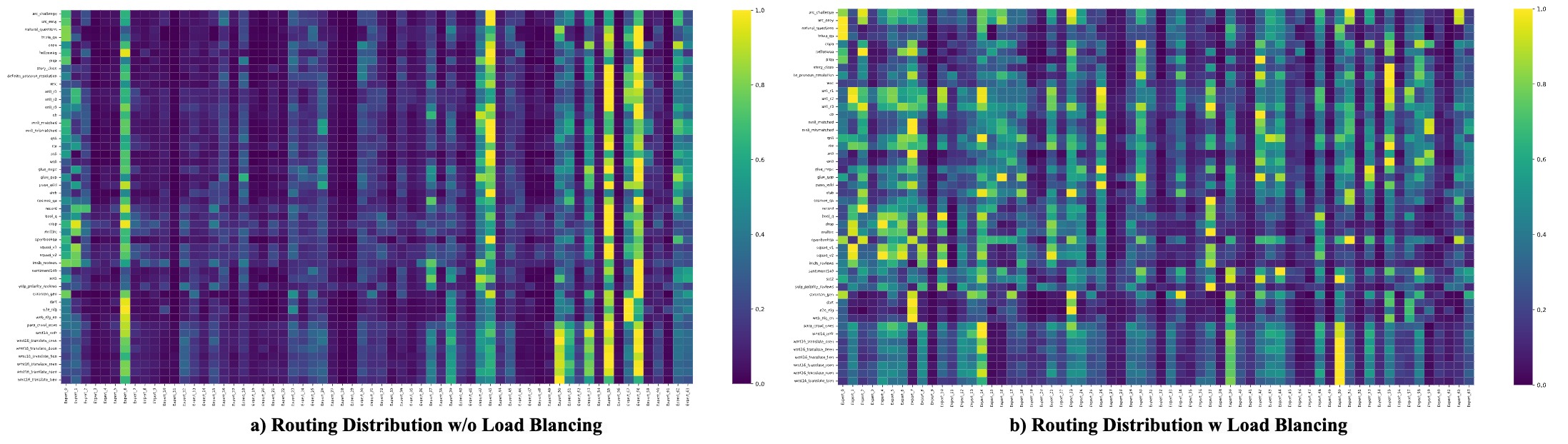}
    \caption{Visualization of Routing Distribution showcasing the impact of the Load Balancing strategy.}
    \label{fig:ablation_routing_vis}
\end{figure*}

To further investigate the role of the load-balancing strategy, we visualized the routing distribution under two conditions: with and without load balancing in Fig.\ref{fig:ablation_routing_vis}. The visualization clearly demonstrates that without load balancing (left), the routing is highly concentrated on a few experts, leading to an uneven workload distribution. In contrast, when the load balancing strategy is applied (right), the routing becomes more evenly distributed across the experts, ensuring a better balance in workload. This highlights the effectiveness of the load balancing mechanism in promoting fair utilization of all available experts.

\section{Comparison of Parameter Counts Across Methods}
\label{app:para_count}
\begin{figure*}[h!]
    \centering
\includegraphics[width=.8\linewidth]{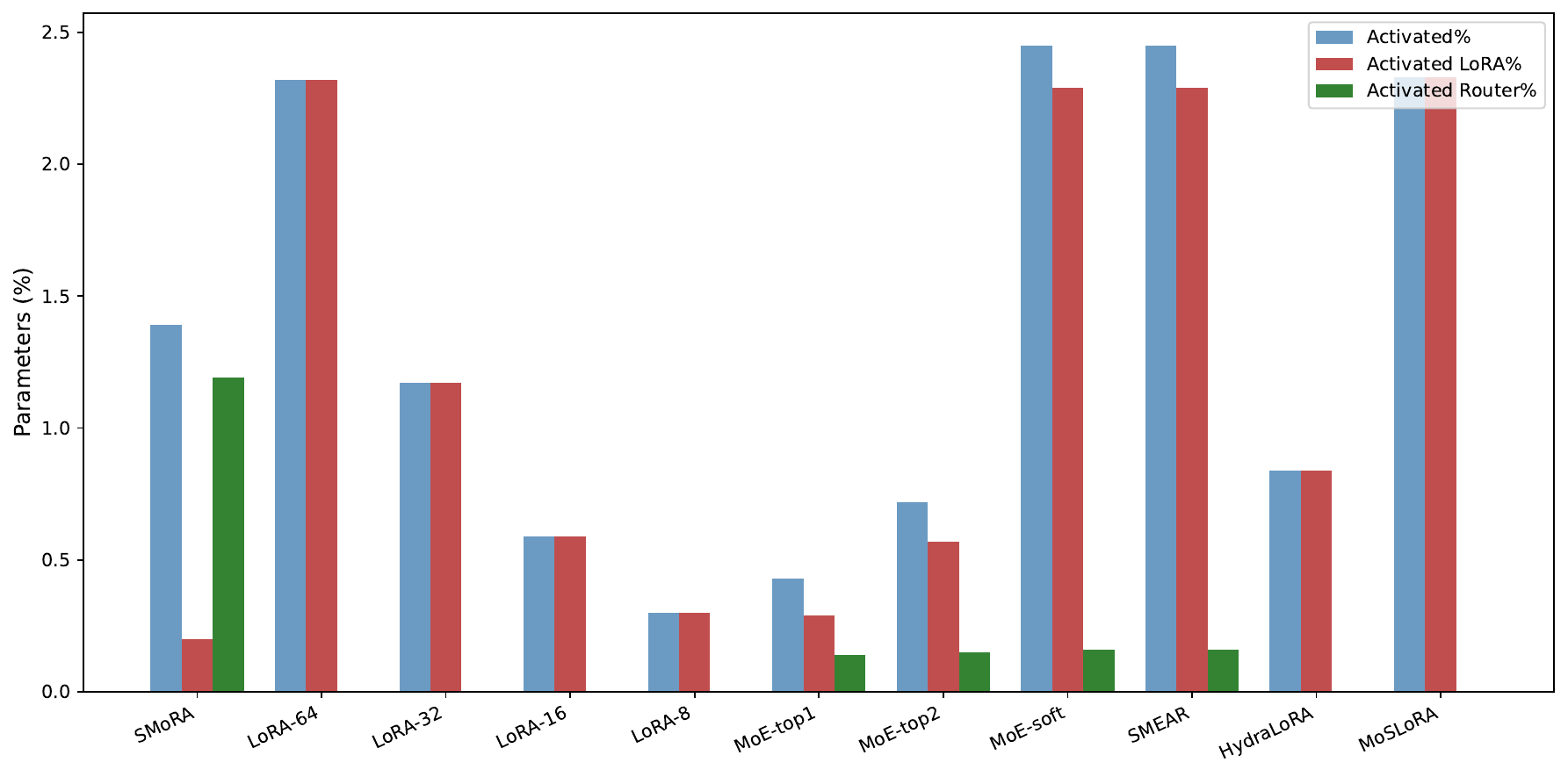}
    \caption{Parameter Counts Across Methods.}
\label{fig:para_cont}
\end{figure*}

In Fig.\ref{fig:para_cont}, we analyze the number of activated parameters across different methods to assess their effectiveness. Regarding the total activated parameters, SMoRA exhibits a higher router parameter count. While it outperforms dense activation methods such as LoRA-64, MoE-Soft, SMEAR, and MoSLoRA, it also activates more parameters compared to sparse activation methods like MoE-top1 and MoE-top2. This is due to its finer-grained expert decomposition, which increases the router parameter count, leading to a higher overall parameter activation. However, when focusing on the activated LoRA parameters, SMoRA activates significantly fewer parameters than all other methods. Ultimately, SMoRA achieves superior performance compared to all methods, regardless of whether they rely on dense or sparse activation strategies.

\end{document}